\newif\ifarxiv
\arxivtrue 

\ifarxiv
    \documentclass[]{fairmeta}
\else 
    \documentclass[runningheads]{llncs}
    \usepackage[review,year=2024,ID=7230]{eccv}

\fi

 \usepackage{eccvabbrv}

 \usepackage{graphicx}
 \usepackage{booktabs}

\usepackage{graphicx}
\usepackage{amssymb}
\usepackage{booktabs}
\usepackage{colortbl}
\usepackage{amsfonts}
\usepackage{stmaryrd}
\usepackage{dsfont}
\usepackage{xfrac}
\usepackage{mathrsfs}
\usepackage{multirow}
\usepackage{tabularx}
\usepackage{tabulary}
\usepackage{tcolorbox}
\usepackage{setspace}
\usepackage[bb=dsserif]{mathalpha}
\usepackage{makecell}
\usepackage{amsmath}
\usepackage{ifthen}

\definecolor{darkred}{RGB}{139, 0, 0}
\definecolor{darkblue}{RGB}{0, 0, 139}

\def\SP#1{\textsuperscript{#1}}
\def\SB#1{\textsubscript{#1}}
\def\SPSB#1#2{%
\ifthenelse{\equal{#1}{}}{\SB{#2}}{\ifthenelse{\equal{#2}{}}{\SP{#1}}{\rlap{\SP{#1}}\SB{#2}}%
}}

\newcommand{\ra}[1]{\underline{{\bf#1}}}
\newcommand{\rb}[1]{{\bf#1}}
\newcommand{\rc}[1]{\underline{#1}}
\newcommand{\rd}[1]{{#1}}

\definecolor{enc}{rgb}{0.85, 0.95, 1}
\definecolor{llm}{rgb}{0.85, 0.98, 0.80}

\definecolor{fext}{rgb}{1, 0.95, 0.85}
\definecolor{finj}{rgb}{1, 0.90, 0.70}
\definecolor{fmap}{rgb}{1, 0.85, 0.85}
\definecolor{ft}{rgb}{1, 0.75, 0.75}

\usepackage{soul}

\newcommand{\llama}{LLaMA}
\newcommand{\llamatwo}{Llama 2}
\newcommand{\promptfuse}{PromptFuse}
\newcommand{\limber}{LiMBeR}
\newcommand{\epalm}{eP-ALM}

\newcommand{\llamaadapter}{LLaMA-Adapter}

\newcommand{\flamingo}{Flamingo}

\newcommand{\bliptwo}{BLIP-2}
\newcommand{\mapl}{MAPL}
\newcommand{\frozen}{Frozen}
\newcommand{\magma}{MAGMA}

\newcommand{\vladapter}{VL-Adapter}

\newcommand{\QPMapper}{QPMapper}
\newcommand{\resamplerRand}{R-rand}
\newcommand{\resamplerLinear}{R-linear}
\newcommand{\resamplerQsformer}{R-\QPMapper}
\newcommand{\resamplerAvgpool}{R-avgpool}

\newcommand{\sunibase}[2]{DePALM\SPSB{#1}{#2}}
\newcommand{\suniInner}[2]{\sunibase{QP,inner#1}{#2}}
\newcommand{\sunicattn}[2]{\sunibase{c-attn#1}{#2}}
\newcommand{\suniLzero}[2]{\sunibase{}{#2}}

\newcommand{\suniRrand}[2]{\sunibase{R-rand,L0#1}{#2}}
\newcommand{\suniRlinear}[2]{\sunibase{R-linear,L0#1}{#2}}
\newcommand{\suniRqsformer}[2]{\sunibase{R-\QPMapper,L0#1}{#2}}
\newcommand{\suniRavgpool}[2]{\sunibase{R-avgpool,L0#1}{#2}}

\newcommand{\largedata}[1]{{\color{orange}#1}}
\newcommand{\mours}[4]{{#1\SPSB{#2}{#3}#4}}
\newcommand{\moours}[4]{\mours{#1}{#2}{#3}{#4 (our reimpl.)}}

\makeatletter
\renewcommand{\paragraph}{%
  \@startsection{paragraph}{4}%
  {\z@}{1ex \@plus 0ex \@minus .0ex}{-1em}%
  {\normalfont\normalsize\bfseries}%
}
\makeatother

\newcommand{\qualimgwidth}{1.8cm}
\newcommand{\qualimgboxwidth}{2.6cm}

\definecolor{lllgray}{RGB}{245,245,245}
\definecolor{llgray}{RGB}{230,230,230}
\definecolor{llred}{RGB}{248,206,204}
\definecolor{llgreen}{RGB}{213,232,212}

\newtcolorbox{imagebox}{
    enhanced,
    boxsep=2pt,left=0pt,right=0pt,top=0pt,bottom=0pt,
    width=\qualimgboxwidth,
    colback={llgray},
    baseline=3.5mm,
    box align=center,
    boxrule=0.5pt,
    nobeforeafter
}

\newtcolorbox{captiontextbox}{
    enhanced,
    boxsep=2pt,left=0pt,right=0pt,top=0pt,bottom=0pt,
    width=\qualimgboxwidth,
    colback={llgreen},
    baseline=3.5mm,
    box align=center,
    boxrule=0.5pt,
    nobeforeafter,
    sharpish corners
}
\newtcolorbox{captiontextboxwrong}{
    enhanced,
    boxsep=2pt,left=0pt,right=0pt,top=0pt,bottom=0pt,
    width=\qualimgboxwidth,
    colback={llred},
    baseline=3.5mm,
    box align=center,
    boxrule=0.5pt,
    nobeforeafter,
    sharpish corners
}

\newcommand{\questiontext}[1]{
{\fontfamily{cmtt}\selectfont\tiny\centering#1\par}
}

\newcommand{\qualtext}[1]{%
\begin{minipage}{.33\linewidth}\centering
    \begin{captiontextbox}
    \tiny\centering\fontfamily{cmtt}\selectfont#1\par
    \end{captiontextbox}
\end{minipage}%
}

\ifarxiv
\else 
    \usepackage[accsupp]{axessibility}  

    \usepackage[pagebackref,breaklinks,colorlinks,citecolor=eccvblue]{hyperref}

    \usepackage{orcidlink}

    \usepackage[capitalize]{cleveref} 
\fi

\ifarxiv
    \author[1]{Th\'eophane Vallaeys}
    \author[2]{Mustafa Shukor}
    \author[2,3]{Matthieu Cord}
    \author[1]{Jakob Verbeek}
    
    \affiliation[1]{FAIR, Meta}
    \affiliation[2]{Sorbonne University}
    \affiliation[3]{Valeo.ai}
    
    \abstract{
    The abilities of large language models (LLMs) have recently progressed to unprecedented levels, paving the way to novel applications in a wide variety of areas.
In computer vision, LLMs can be used to prime vision-language tasks such image captioning and visual question answering when coupled with pre-trained vision backbones. 
While different approaches have been explored to interface LLMs with ``perceptual backbones'' that process, \eg, visual or audio data, they are often explored for different tasks, different datasets, and using different perceptual backbones and language models, hindering direct comparison of the interfacing mechanisms.
To remedy this lack of comparability between methods,
we present an extensive experimental evaluation of different interfacing mechanisms, across multiple tasks (including image, video, and audio captioning as well as visual question answering), datasets and backbones, paying special attention to low-data settings. 
We find improved performance using existing mechanisms over state-of-the-art results, and identify a new interfacing mechanism that yields (near) optimal results across different tasks, while obtaining  a $4\times$ reduction in training time. 
    }
    
    \date{\today}
    \correspondence{Th\'eophane Vallayes (\email{webalorn@gmail.com}) Mustafa Shukor (\email{mustafa.shukor@sorbonne-universite.fr})
    \vspace{10mm}
    }

\fi

\begin{document}

\title{Improved Baselines for Data-efficient\\Perceptual Augmentation of LLMs} 

\maketitle

\begin{figure}[t]
    \centering
    \begin{minipage}{.58\linewidth}
    \centering
    \begin{subfigure}[b]{\textwidth}
       \includegraphics[width=\linewidth]{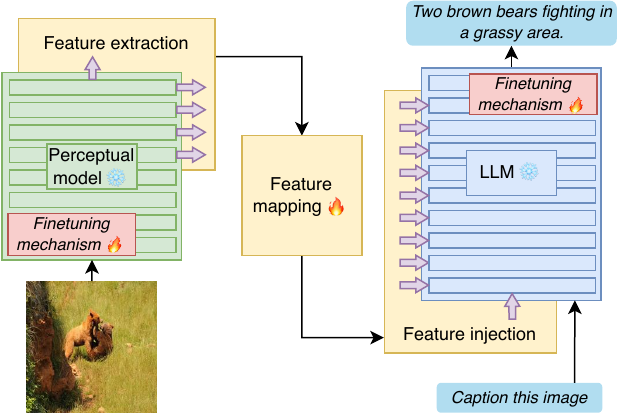}
       \caption{Our unified framework for perceptual augmentation of LLMs.
   }
        \end{subfigure}
    \end{minipage}%
    \hfill
    \begin{minipage}{.4\linewidth}
    \centering
        \begin{subfigure}[b]{\textwidth}
\includegraphics[width=\linewidth]{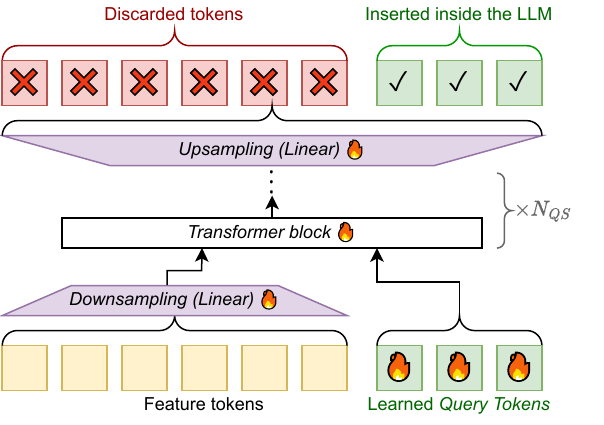}
   \caption{Architecture of the feature mapping (\QPMapper{}) in \suniLzero{}{}.
   }
    \end{subfigure}
        
    \end{minipage}%
   \caption{(left) Unified view of data-efficient approaches for perceptual augmentation of LLMs. Existing approaches can be characterized by four configurable  blocks: feature extraction, mapping, and injection, as well as  fine-tuning mechanisms. (right) Architecture of the feature mapping (\QPMapper{}) used by our model (DePALM): it aggregates tokens from the encoder before injecting them in the LLM, which results in significant savings in compute.
   }
    \label{fig:general-framework-2}
\end{figure}

\section{Introduction}
\label{sec:intro}

The advent of large language models (LLMs) has brought unprecedented capabilities in the understanding and production of natural language~\cite{opt,llama,llama2,gpt3,gpt-j}. 
These models can be leveraged to provide a natural user interface in a wide variety of applications, including text-based generation of  images, video and audio~\cite{ramesh22dalle2,liu2023audioldm,ho2022imagen}, using external tools~\cite{schick2023toolformer} and make  models talk to each other~\cite{zeng2022socratic}.

Currently, state-of-the-art models in image captioning~\cite{Labb2023CoNeTTEAE,Hu2022ExpansionNetVB} and visual question-answering (VQA)~\cite{chen2022pali,He2023VLABEV} mostly consist of task-specific, end-to-end trained models. To build more general models, beyond a single task or dataset, several works leverage the generalization capabilities of pre-trained LLMs, coupled with visual encoders~\cite{flamingo,chen2022pali,Wu2023NExTGPTAM,blip2,unival}. 
Such approaches rely on end-to-end training of very large numbers of parameters,  \eg 10B in Flamingo~\cite{flamingo}, requiring very large datasets, \eg up to several billions of examples in~\cite{chen2022pali}.
Recently, a significant efforts have been focused on building more powerful multimodal models \cite{liu2023improvedllava15,bai2023qwenvl,liu2024visualinstructblip,lin2023sphinx,chen2023minigptv2}. Yet, they  still involve  costly training stages, such as multimodal pretraining and multitask instruction tuning. These models are interesting when millions of training samples are available and there are no constraint on compute efficiency, and the goal is to have generalist models with good performance on many tasks.
An  interesting alternative line of research has emerged that studies data and parameter efficient methods ~\cite{epalm,mapl,llamaadapterv2,frozen,clipcap,vl-adapter,limber} to address multimodal tasks. 
These approaches focus on adapting pre-trained and frozen LLMs, by training modules with few  parameters, on limited training sets. This line of research is complementary to the former, and aims to maximize  performance on a specific  task (\eg VQA), within few hours of training on a single machine. This becomes even more important in case of data scarcity where we lack big datasets with millions of training samples. 

Improvements in parameter-efficient approaches span several axes, such as the LLMs and perceptual encoders used~\cite{limber, clipcap, epalm, llamaadapter}, the perceptual feature extraction/injection mechanism~\cite{epalm,limber,llamaadapter}, and the cross-modal mapping module~\cite{mapl,frozen,limber}. 
This variety of design choices prevents a fair and comprehensive comparison between existing approaches, and hinders the understanding of the main factors driving their success. In addition, most of these approaches focus on parameter-efficiency, with little focus on data-efficiency~\cite{epalm,mapl}, while we argue that the latter is a more important aspect together with compute-efficiency.

More than proposing novel approaches to couple LLMs with perceptual backbones, we believe it is important to have a unified understanding  and proper comparison between existing methods. To this end, we propose a unified framework to comprehensively study previous approaches~\Cref{fig:general-framework-2}. 
Our framework allows a fair and systematic comparison along designs of several blocks:  feature extraction (\emph{e.g.} which visual features to consider), feature mapping (\emph{e.g.} how to project the extracted features in the LLM textual space) and feature injection (\emph{e.g.} where to inject the projected features). 
We consider the impact of the choice of LLM and perceptual backbones, and carefully and fairly tune  hyperparameters. This by itself already improves over previously reported results. 
The systematic characterization of existing approaches naturally leads us to define and evaluate alternative approaches.
We find that one of these approaches emerges as the overall best, which we dub \suniLzero{}{}, leading to (near) optimal results across different datasets and tasks. 
Our approach consistently and significantly improves over earlier data and parameter efficient approaches, and in some cases also outperforms few-shot performance of large-scale state-of-the-art models that train billions of parameters on massive datasets.

To summarize, our contributions are as follows:
\begin{itemize}
  \setlength\itemsep{0em}
  
\item We present the first systematic experimental study of methods to interface  perceptual backbones with LLMs, using the same tasks, datasets, and underlying backbone networks. 

\item For all considered tasks, we find improvements over previous state-of-the-art data and parameter efficient methods by careful setting of training hyperparameters and architectural choices. 

\item We identify a new mechanism,  \suniLzero{}{}, to interface LLMs with perceptual backbones based on token pooling, which obtains near optimal  results, while being up to 4$\times$ faster to train than the closest competitor (training in less than 1.5h on a single machine for a typical dataset). 

\end{itemize}

\section{Related work}

\begin{table}[ht] 
\small
\centering	

\caption{
    Overview of different architectures from the literature, as well as from this work (DePALM models). 
    The LLM adaptation mechanisms consist of four fundamental components: feature extraction, feature mapping, feature injection, and a fine-tuning mechanism.
    The last column shows the number of trainable parameters, as reported by papers, or with the \llama+CLIP-L setting in our models.
    \textit{\largedata{Methods in orange} leverage pre-training on large amounts of data, or cross-dataset training. Others have at least one version trained on a single dataset, which is the setting we consider.}
    }
    
\setlength\tabcolsep{4pt}
\resizebox{\linewidth}{!}{%
\begin{tabular}{c|cc|cccc|c}
\toprule
\multirow{2}{*}{\textbf{Method}} & \multicolumn{2}{c|}{\textbf{Backbones}} & \multicolumn{4}{c|}{\textbf{Adaptation mechanism}}  & \textbf{\# Tr.} \\ \cmidrule{2-3} \cmidrule{4-7} 
& \textbf{LLMs} & \textbf{Perceptual Enc.} & \textbf{Feature extraction}  & \textbf{Feature mapping} & \textbf{Feature injection} & \textbf{Fine-tuning mechanisms}  & \textbf{params.} \\

\midrule
\largedata{\flamingo{}}~\cite{flamingo}
    & \makecell{Chinchilla \cite{Hoffmann2022TrainingCL}} & \makecell{NFNet \cite{NFResnet}}
    & \makecell{Tokens from last layer}
    & \makecell{Perceiver Resampler \\ (Transformer)}
    & \makecell{GATED XATTN-DENSE \\ (Cross-attention)}
    & \makecell{--}
    & 10B \\
\midrule
\largedata{\bliptwo{}}~\cite{blip2}
    & \makecell{OPT \cite{opt}, FlanT5 \cite{flant5}} & \makecell{CLIP \cite{clip}}
    & \makecell{Tokens from last layer}
    & \makecell{Q-Former}
    & \makecell{1st layer token injection}
    & \makecell{--} 
    & 1.2B \\
\midrule
\largedata{\magma{}}~\cite{magma}
    & \makecell{GPT-J 6B \cite{gpt-j} } & \makecell{ CLIP \cite{clip} / NFNet \cite{NFResnet}}
    & \makecell{Tokens from last layer}
    & \makecell{MLP}
    & \makecell{1st layer token injection}
    & \makecell{fine-tuning of \\ perceptual model} 
    & 243M \\
\midrule
\mapl{}~\cite{mapl}
    & \makecell{GPT-J 6B \cite{gpt-j}} & \makecell{CLIP-L \cite{clip}}
    & \makecell{Tokens from last layer}
    & \makecell{\QPMapper{} \\ ($d_\text{embed}$=256, 4 layers)}
    & \makecell{1st layer token injection}
    & \makecell{--} 
    & 3.4M \\
\midrule
\promptfuse{}~\cite{promptfuse}
    & \makecell{BART \cite{bart}} & \makecell{ ViT \cite{dosovitskiy21iclr}}
    & \makecell{Tokens from last layer}
    & \makecell{\textit{nothing}}
    & \makecell{--}
    & \makecell{prompt tuning} 
    & 15K \\
\midrule
\largedata{\limber{}}~\cite{limber}
    & \makecell{GTP-J 6B \cite{gpt-j} } & \makecell{CLIP \cite{clip}}
    & \makecell{Tokens from last layer}
    & \makecell{Linear projection}
    & \makecell{1st layer token injection}
    & \makecell{--} 
    & 12.5M \\
\midrule
\epalm{}~\cite{epalm}
    & \makecell{OPT-2.7B/6.7B \cite{opt} } & \makecell{ ViT \cite{steiner2021augreg}, AST \cite{ast}, \\ TimeSformer \cite{timesformer}}
    & \makecell{CLS tokens from \\ $n$ last layers}
    & \makecell{(Shared) \\ linear projection}
    & \makecell{Token injection in \\ intermediate layers}
    & \makecell{prompt tuning} 
    & 4.2M \\
\midrule
\llamaadapter{}~\cite{llamaadapter,llamaadapterv2}
    & \makecell{\llama \cite{llama} } & \makecell{ CLIP \cite{clip}}
    & \makecell{Tokens from last layer}
    & \makecell{Linear projection}
    & \makecell{Token injection in \\ intermediate layers}
    & \makecell{inner-layer prompt tuning, \\ bias tuning, norm tuning} 
    & 14M \\
\midrule
\frozen{}~\cite{frozen}
    & \makecell{GPT-like \cite{gpt2}} & \makecell{NFNet \cite{NFResnet}}
    & \makecell{Pooled output tokens}
    & \makecell{\textit{nothing}}
    & \makecell{1st layer token injection}
    & \makecell{Fine-tune the NFNet} 
    & 40.3M \\
\midrule
ClipCap~\cite{clipcap}
    & \makecell{GPT-2\cite{gpt2}} & \makecell{CLIP \cite{clip}}
    & \makecell{Tokens from last layer}
    & \makecell{Transformer}
    & \makecell{1st layer token injection}
    & \makecell{--} 
    & 43M \\
\midrule
VL-Adapter~\cite{vl-adapter}
    & \makecell{BART \cite{bart}, T5 \cite{T5} } & \makecell{ CLIP \cite{clip}}
    & \makecell{Tokens from last layer}
    & \makecell{Linear projection}
    & \makecell{1st layer token injection}
    & \makecell{Adapters} 
    & 5.8M \\
\midrule
\largedata{AnyMAL}~\cite{anymal}
    & \makecell{\llamatwo-70B-chat \cite{llama2}, } & \makecell{CLIP \cite{clip}, CLAP \cite{clap}}
    & \makecell{Tokens from last layer}
    & \makecell{Perceiver Resampler,\\or linear projection}
    & \makecell{1st layer token injection}
    & \makecell{LoRA \cite{lora}} 
    & -- \\

\midrule
\midrule

\suniInner{}{}
    & \multirow{7}{*}{\makecell{OPT-6.7B \cite{opt},\\ \llama{} \cite{llama} }} & \multirow{7}{*}{\makecell{CLIP-L \cite{clip}, \\ DINOv2 \cite{dinov2}, \\
    MAViL \cite{mavil} \\
    TimeSformer \cite{timesformer}}}  
    & \multicolumn{1}{c|}{\makecell{Tokens from \\ $n$ last layers}}
    & \multicolumn{1}{c|}{\multirow{3}{*}{\makecell{\QPMapper{}}}}
    & \multicolumn{1}{c}{\makecell{Token injection in \\ intermediate layers}}
    & \multicolumn{1}{|c|}{\multirow{7}{*}{prompt tuning}} 
    & \multicolumn{1}{c}{18.1M} \\
\cmidrule{1-1}\cmidrule{4-4}\cmidrule{6-6}\cmidrule{8-8}
\suniLzero{}{}
    & 
    &
    & \multicolumn{1}{c|}{\multirow{4}{*}{Tokens from last layer}}
    & 
    & \multicolumn{1}{|c|}{\multirow{4}{*}{\makecell{1st layer token injection}}}
    & 
    & \multicolumn{1}{c}{17.9M} \\
\cmidrule{1-1}\cmidrule{5-5}\cmidrule{8-8}
\makecell{\suniRrand{}{}, \suniRlinear{}{}, \\ \suniRqsformer{}{}, \suniRavgpool{}{}}
    & 
    & 
    &
    & \multicolumn{1}{|c|}{\makecell{Linear projection \\ + Resampler}}
    &
    & 
    & \multicolumn{1}{c}{\makecell{21M, 88M\\18M, 21M}} \\
\cmidrule{1-1}\cmidrule{4-6}\cmidrule{8-8}
\sunicattn{}{}
    & 
    &
    & \multicolumn{1}{c|}{\makecell{Tokens from \\ $n$ last layers}}
    & \multicolumn{1}{c|}{\makecell{Projection \\ + Small Transformer}}
    & \multicolumn{1}{c|}{\makecell{Gated cross-attention}}
    & 
    & \multicolumn{1}{c}{17.9M} \\

\bottomrule
\end{tabular}
}
\label{tab:architectures}
\end{table}

\paragraph{Multimodal models.}
In recent years there has been a significant interest in  multimodal models, and in particular in vision-language pre-training, see \eg~\cite{tan2019lxmert, uniter, ALBEF, shukor2022efficientvicha, dou2022empiricalmeter, blip1,clip}. 
These models can be subsequently fine-tuned to address a range of tasks, such as visual question answering (VQA) and image captioning. 
The advent of large language models (LLMs)~\cite{gpt3, hoffmann2022trainingchinchilla, opt, llama, chowdhery2022palm} has triggered another line of work on large-scale multimodal training built on top of LLMs.
The typical approach is fine-tune a pre-trained language model on large multimodal datasets~\cite{chen2022pali, chen2023pali}. 

Due to the large computational cost to train these approaches, especially with LLMs on the scale of  billions of parameters, other approaches keep the LLM part of the model frozen, and only train additional parameters to solve multimodal tasks~\cite{flamingo, blip2}. 
Although the predominant focus of research revolves around image-text tasks, the adaptability of these approaches to other modalities, including video and audio, has recently been demonstrated in a straightforward manner~\cite{epalm,Han2023ImageBindLLMMI,unival,Lyu2023MacawLLMML,unival,anymal,Wu2023NExTGPTAM}.
Nonetheless, such endeavors still necessitate the training of a substantial number of parameters, \eg 10B parameters in Flamingo~\cite{flamingo}, on billion-scale  multimodal training sets~\cite{chen2023pali}.

\paragraph{Efficient adaptation of unimodal models.}
In contrast to the  paradigm of large-scale end-to-end multimodal training, another line of work considers efficient adaptation of pre-trained unimodal models.  
Methods such as MAGMA~\cite{magma}, Frozen~\cite{frozen} and ClipCap~\cite{clipcap} tackle vision-language tasks by training the visual encoder~\cite{frozen,magma} or additional adapters~\cite{magma} to leverage a  pre-trained language model. 
Other approaches train smaller number of parameters by keeping all pre-trained models fixed, and train a linear layer~\cite{limber} or a small transformer mapping network~\cite{mapl}. 
Nevertheless, these methods rely on multimodal visual encoders such as CLIP~\cite{clip}, or inject a substantial number of visual tokens in the language model, which  reduces inference speed. 
Recently, several approaches ~\cite{epalm,llamaadapter, koh2023groundingfromage} have explored the use of simple linear layers to transform features and inject them in LLMs, some of them  even use only-unimodal encoders, across image/video/audio modalities, see \eg~\cite{epalm}. 
While each of these approaches show good performances within its own specific experimental setup, it is difficult to compare them due to the differences in  the considered tasks and  datasets.

\section{Unified framework}

Even though different approaches leverage LLMs for multimodal tasks, it remains challenging to discern the specific components responsible for the superiority of one method over another. 
In response to this challenge, we structure previous work in a comprehensive and unified framework, as depicted in \Cref{fig:general-framework-2}, enabling a systematic and fair comparison of various existing approaches. 
Within this framework, the process of adapting LLMs for multimodal tasks boils down to make different design choices. 
These include the choice of the LLM and perceptual backbone models, which we discuss in \Cref{sec:backbones}, and the adaptation mechanism, which we discuss in \Cref{subsec:adaptation_mechanisms}. The latter consists of a feature extraction, feature mapping and feature injection mechanism, and finally a fine-tuning mechanism.
Different design choices lead to different existing or new approaches, as exemplified for a number of methods from the literature in \Cref{tab:architectures}. 

\subsection{Backbone models}
\label{sec:backbones}

\paragraph{Language models.} Despite a non-negligible effort in encoder-decoder LLMs, the NLP community has mostly converged to decoder-only LLMs for very large scales. Most powerful LLMs come with different models sizes, the best choices in terms of the trade-off between performance and efficiency are usually models with in the order of  7B parameters. 
To solve multimodal tasks, these LLMs can be fully or partly finetuned, or completely frozen. Here we focus on frozen, decoder-only LLM with 7B parameters, such as OPT~\cite{opt} and \llama{}~\cite{llama}.  We also experiment with intruction-tuned LLMs such as Vicuna~\cite{vicuna} and LlaMA-2-chat{}~\cite{llama2}. 

\paragraph{Perceptual encoders.}
Encoders are chosen depending on the modality, and they differ mainly in the architecture (\eg CNNs or Transformers) and training paradigm (\eg class-label supervised, text supervised, or self supervised). 
In our experiments, we focus on transformer-based encoders for their strong performance when pre-trained on large-scale datasets~\cite{dosovitskiy21iclr}. 
We experiment with CLIP~\cite{clip} for image-text tasks, which has been pre-trained from text-aligned data.
We also experiment with models pre-trained in an unimodal manner such as TimeSformer~\cite{timesformer} for videos, and self-supervised ones such as DINOv2~\cite{dinov2} for images, and MAViL~\cite{mavil} for audio.

\subsection{Adaptation mechanisms}
\label{subsec:adaptation_mechanisms}
To couple the perceptual backbone with the LLM, perceptual tokens are first extracted from the perceptual backbone, transformed via a mapping network, and then injected in the LLM. To further improve the adaptation, different fine-tuning mechanisms can be adopted. 
An overview of the different designs of these components is given in columns four to seven in \Cref{tab:architectures}.
Below, we  discuss each of them in detail.

\paragraph{Feature extraction.}
\label{sec:feat_extraction}
In transformer-based perceptual encoders, features take the form of ``tokens'' that  correspond to specific parts of the input, \eg an image patch. 
Some transformer models also include a special ``class token'',  denoted as \textsc{CLS}, which is not tied to a specific part of the input; it interacts with all the input-tied tokens, and encodes global information that can, \eg, be used to classify an image.
These tokens can be extracted from any layer of the encoder. 
We consider two design choices; (i) where to extract the tokens:  from the last encoder layer only, or from the last $k$ layers, and (ii) which tokens to extract: all of them, or only the \textsc{CLS} token.

\paragraph{Feature mapping.}
\label{sec:feat_mapping}
To render the encoder features compatible with the internal features of the LLM, we apply a  mapping which can take different forms.

\paragraph{1) Linear projection.} The simplest approach is to use a single linear layer that  projects the extracted visual tokens to have the same LLM hidden state dimension. 
If multiple tokens are extracted, each of them is projected independently with the same linear layer.

\paragraph{2) Query pooling mapper.} 
Typical perceptual backbones use in the order of hundreds of internal tokens, \eg 256 tokens organized in a  $16\times 16$ grid for images.
The training and inference cost directly depend on the number of tokens that are extracted from the encoder and injected in the LLM. 
Even if the LLM in principle needs to process only short text sequences for the task at hand, such as in visual question answering, injecting hundreds of perceptual tokens in the LLM makes it computationally demanding.
We design a {\bf \QPMapper{} } block to aggregate the tokens  extracted from the encoder into a smaller set. 
The input feature tokens are projected and concatenated to a sequence of learnable query tokens (hence the name), and only the outputs corresponding to the query tokens are kept, upsampled to the LLM dimension, and normalized using the RMSNorm~\cite{rms-norm}.
This architecture is inspired by several previous works~\cite{blip2, douillard2022dytox,touvron2021goingcait} using query or class tokens to compute an aggregate representation of the input. In our case, however, we use multiples rather than a single global token.
It is also similar to the mapping network of \mapl~\cite{mapl}, with a lower number of layers and higher dimensionality, and its main benefit is to limit the number of tokens passed to the LLM.
Our \QPMapper{} consists of a small sequence of $N_{QS}$ transformer layers, wrapped by linear dimension downsampling and upsampling  projections to a $d_\text{embed}$-dimensional internal features, allowing to control the number of trainable  parameters in this block.
See \Cref{fig:general-framework-2} (right panel) for an illustration of QPMapper architecture.

\paragraph{3) Token resamplers.}
We consider several other alternatives to reduce the number of tokens, which are inspired from  pooling blocks  in CNNs.
For example, average-pooling and max-pooling aggregate features over a small patch of, typically, $2\times2$ features. 
In our experiments, we explore the following resamplers to reduce the number of tokens:
\begin{itemize}
    \item \resamplerAvgpool{}:  tokens in a patch are averaged, which is equivalent to an average pooling layer on the input grid.
    \item \resamplerLinear{}: tokens in a patch   are concatenated and then linearly projected, 
     which is equivalent to a strided convolution on the input grid.
    \item \resamplerQsformer{}:  tokens in a patch   are passed through a \QPMapper{} with a single query token, with parameters of the \QPMapper{} shared across patches.
    \item \resamplerRand{}: a random subset of tokens, \eg 50\%, is selected during training. During evaluation, we keep all tokens.
\end{itemize}

\paragraph{4) Cross-attention.}
Prepending tokens to the textual tokens inside an LLM significantly increase the inference complexity. 
Rather than reducing the number of prepended tokens, we consider a parameter-efficient cross-attention module that is inserted in the LLM and allows it to access the tokens of the perceptual encoder,  inspired by Flamingo~\cite{flamingo}.  
Specifically, the perceptual and textual tokens are projected to a smaller hidden dimension $d_\text{embed}$. In this latent space, a typical cross-attention block is applied. The textual tokens are considered as query, and the perceptual ones as keys and values. Finally, the output is upsampled to the LLM dimension and added back to the initial textual tokens using a tanh-gated residual connection. These modules are inserted throughout the second half of the LLM, in-between the LLM transformer modules.

\paragraph{Feature injection.}
\label{sec:feat_injection}

As for the injection of tokens in the LLM, there are two choices to make: (i) how to inject, and (ii) where to inject these tokens. 
Regarding (i),  we can  prepend the tokens to the textual tokens and then interact with the text in the LLM self-attention layers, or we inject them via a cross-attention mechanism. 
Regarding (ii), tokens are either injected in the LLM input layer, or in intermediate ones.  
When prepending the tokens to textual tokens in the input layer, they propagate up until the last layer. 
When injecting in intermediate layers, they are only kept for a single attention block and discarded afterward, and possibly replaced by the same set or another set of tokens in the next transformer block. 
In this last case, if the tokens were extracted from $k$ levels of the perceptual model, and inserted in to $n$ LLM layers, then each sequence $i=1,\dots, k$ of perceptual features is inserted into the LLM blocks $\lfloor in/k \rfloor, \dots,  \lfloor (i+1)n/k -1 \rfloor$.

\paragraph{Finetuning mechanisms}

While keeping the LLM frozen is most efficient, parameter-efficient fine-tuning techniques 
can be used to further boost performance~\cite{ding2022delta}.
In our  experiments, we consider prompt-tuning and bias-tuning, which we  detail in the supplementary material.

\section{Experiments}

Below, we present our experimental setup in \Cref{sec:setup},
followed by the results in \Cref{sec:results}.

\subsection{Experimental setup}
\label{sec:setup}

\paragraph{Datasets and metrics.}
The datasets used in our experiments are listed in \Cref{tab:datasets}.
For all datasets we use standard splits, except for COCO and VQAv2 where we use the commonly used Karpathy splits~\cite{Karpathy_2015_CVPR}. 
To study limited data settings, we consider OKVQA and MSRVTT, and also experiment with COCO and VQAv2 using 1\% of the training data.
We evaluate using the standard metric of each benchmark. 
Specifically, for both image and video captioning, we use CIDEr~\cite{vedantam2015cider}, and for VQA tasks, we use the official VQAv2 accuracy metric on the test and/or validation set.  Audio captioning is evaluated with SPIDER\cite{spider}. We add other standard metrics (BLEU~\cite{papineni2002bleu}, METEOR~\cite{meteor}, SPICE~\cite{spice}) in the supplementary material.

\paragraph{Baselines.}
To ensure fair comparison of the different interfacing mechanisms, we re-implement several parameter-efficient approaches: \limber{}~\cite{limber}, \mapl{}~\cite{mapl} and \epalm{}~\cite{epalm}. 
We selected \limber{} as this is the simplest method, used in a number of other works, and the other two as their original paper also report results on the data-efficient setting.
We found the other models either redundant (VL-adapter\cite{vl-adapter} is the same as \limber{} with additional fine-tuning, which is not our main focus), non-parameter or non-data efficient (\frozen{}~\cite{frozen}, \bliptwo{}~\cite{blip2}) or designed for another setting (\llamaadapter{}~\cite{llamaadapter} was conceived for instruction fine-tuning first).
We refer to these as \moours{\limber}{(all)}{}{}, \moours{\mapl}{}{}{} and 
 \moours{\epalm}{}{}{}, 
and note that we change the backbones from the original papers to be all the same, for proper comparisons. 
We also use a variant of \limber{} from~\cite{epalm}, which we name \moours{\limber}{(1)}{}{},
where only the \textsc{CLS} token is injected in the LLM.

\begin{table}[t]
\centering	
\caption{
    Datasets used in our experiments, listing the modality type, task, and size of the training set. We also list the   LLM and perceptual backbone used by default for each dataset.
}
\label{tab:datasets}

\setlength\tabcolsep{4pt}
\resizebox{0.7\linewidth}{!}{%
\small
\begin{tabular}{lllr|ll}
\toprule
Dataset & Type & Task  & Size & LLM & Backbone \\
\midrule
COCO~\cite{coco} & Image & Captioning & 82K & \llama{}-7B & EVA-CLIP-L  \\
TextCaps~\cite{textcaps} & Image & Captioning & 21K & \llama{}-7B & CLIP-ViT-L  \\
VQAv2~\cite{vqav2} & Image & Question Ans. & 605K & OPT-6.7B & CLIP-ViT-L  \\
TextVQA~\cite{textvqa} & Image & Question Ans. & 34K & OPT-6.7B & CLIP-ViT-L  \\
AOKVQA~\cite{aokvqa} & Image & Question Ans. & 17K & OPT-6.7B & CLIP-ViT-L  \\
OKVQA~\cite{okvqa} & Image & Question Ans. & 9K & OPT-6.7B & CLIP-ViT-L  \\
AudioCaps~\cite{audiocaps} & Audio & Captioning & 49K & OPT-6.7B & MAViL  \\
MSRVTT~\cite{msrvtt} & Video & Captioning & 7K &\llama{}-7B  & TimeSformer  \\
\bottomrule
\end{tabular}
}
\end{table}

\paragraph{Our models.} Based on our unified framework, we explore seven novel interfacing mechanisms, summarized in \Cref{tab:architectures}: \sunibase{QP,L0}{} (that we refer to as \suniLzero{}{}), \suniInner{}{}, \sunicattn{}{}, \suniRrand{}{}, \suniRlinear{}{}, \suniRqsformer{}{} and \suniRavgpool{}{}. 
To get a good trade-off between performance and efficiency, we include different pooling strategies to reduce the number of perceptual tokens, contrary to prior work that either used a single token~\cite{epalm} or all tokens~\cite{limber}. 
Most of these variants extract features from the last perceptual encoder layer and inject the mapped features in the first LLM layer. 
In addition, we also explore models that consider intermediate layers as in~\cite{epalm}. 
In terms of fine-tuning mechanism, we consider prompt tuning and bias-tuning, due to its effectiveness in previous work~\cite{epalm,promptfuse}, and leave the LLM and perceptual backbone frozen. 
In the appendix we report additional experiments regarding different finetuning approaches.

Please refer to the supplementary material for further architectural detail of the  baselines and our models.

\paragraph{Implementation details.} 
For fair comparison between different approaches, we use a unified training setup.
For each dataset, we use the same LLM and perceptual encoders for all methods, as listed in \Cref{tab:datasets}.
Models are trained directly on downstream tasks, without any pre-training, and using the standard cross-entropy loss; except for the LLM and perceptual backbone which are pre-trained and frozen.
We use random perturbations for data-augmentation, using the same procedure as in~\cite{ALBEF} for images.
We train with the AdamW~\cite{Loshchilov2017DecoupledWD} optimizer and the cosine learning rate scheduler~\cite{cosine_scheduler}.
For each experiment, we conduct five different runs with different random seeds, unless specified otherwise, each run being executed on a single machine equiped with eight V100 GPUs.
We report the mean performance metrics in the main paper, and refer to the supplementary material for the standard deviations. 
Further  implementation details can also be found in the supplementary material.

\begin{table}[t]

\centering	
\caption{
    Comparison of our implementation of baselines with results reported by the original papers. Results  averaged  over 5 runs. 
    The best result per column are marked  in bold.
    $\dagger$: The published results for LiMBeR use   the standard split and a 4-shot evaluation using a model trained on a larger dataset, which do not correspond directly to our setting.
    $\ddagger$: results using 8 shots, after training on the target dataset only.
    }
\label{tab:baselines_comp}

\setlength\tabcolsep{4pt}

\resizebox{0.8\linewidth}{!}{%
\begin{tabular}{lccccc}
\toprule	 	
\multirow{2}{*}{Method}
    & COCO $\uparrow$
    & COCO (1\%) $\uparrow$
    & VQAv2 $\uparrow$
    & VQAv2 (1\%) $\uparrow$
    & OKVQA $\uparrow$
    \\
\cmidrule(lr{8pt}){2-2} \cmidrule(lr{8pt}){3-3} \cmidrule(lr{8pt}){4-4}  \cmidrule(lr{8pt}){5-5} \cmidrule(lr{8pt}){6-6}
    & CIDEr
    & CIDEr
    & Val
    & Val
    & Val
    \\
\midrule
\limber{} (4-shot)~\cite{limber}
    & --       
    & --       
    & 39.2$^\dagger$       
    & --       
    & --            
    \\
\mapl{}~\cite{mapl}
    & 125.2       
    & 65.9       
    & 43.5       
    & 37.7       
    & 18.7 / 31.6$^\ddagger$             
    \\
\epalm{}~\cite{epalm}
    & 111.6       
    & --       
    & 54.9     
    & 41.9$^\dagger$       
    & --            
    \\

\midrule
\moours{\limber}{(all)}{}{}
    & \bf 136.3        
    & \bf 83.7        
    & \bf 73.4        
    & \bf 48.0        
    & \bf 36.2             
    \\
\moours{\mapl}{}{}{}
    & 126.1        
    & 69.2        
    & 67.1         
    & 45.9         
    & \bf 36.2             
    \\
\moours{\epalm}{}{}{}
    & 115.3       
    & 64.7        
    & 59.3        
    & 41.4        
    & 23.5        
    \\

\bottomrule
\end{tabular}
}
\end{table}

\subsection{Main experimental results}
\label{sec:results}

\paragraph{Improved baseline performances.} 
We start with our  reproductions of existing parameter and data-efficient baselines. 
\Cref{tab:baselines_comp} shows a comparison between the scores we obtained and those reported in the original papers. 
We improve the existing baselines by large margins across all metrics. 
This comes mainly from using better backbones (\eg, LlaMA and CLIP), and using a thorough hyperparameter search for the training algorithm. 
We conducted this hyperparameter search independently for each experiment, on the learning rate and gradient clipping, using a grid search over a set of values we found to work particularly well for a set of diverse models on our task.
With our implementation, \limber\SP{(all)} achieves the best performance across the board. 
However, \limber\SP{(all)} is computationally more expensive as it dramatically increases the length of the sequence processed by the LLM as by passing  all (typically 256) perceptual tokens to the LLM, compared to passing couple of tokens in \mapl{},  or just one in \epalm{}.

\begin{table}[t]

\centering	
\caption{
    Comparison of our proposed DePALM architectures and our baseline re-implementations, after training on 100\% or 1\% on each datasets. We highlight the \ra{first}, \rb{second} and \rc{third} best results. All results are averaged on 5 runs. 
    We show the \textbf{training time} on AudioCaps, the average rank and average normalized score of each method across benchmarks. For these last two values, we add the rank over all our models.
    $ ^\blacksquare$: tokens are first averaged across time to prevent memory errors.
    $ ^\blacklozenge$: incomplete data due to unstable training. 
}
\setlength\tabcolsep{4pt}
\resizebox{\linewidth}{!}{%
\begin{tabular}{lccccccccccccc}
\toprule	 	
\multirow{2}{*}{Method}
    & COCO
    & COCO (1\%)
    & TextCaps
    & VQAv2
    & VQAv2 (1\%)
    & TextVQA
    & OKVQA
    & AOKVQA
    & AudioCaps
    & MSRVTT
    & \multirow{2}{*}{\makecell{Train\\time $\downarrow$}}
    & \multicolumn{2}{c}{Average}
    \\
\cmidrule(lr{8pt}){2-2} \cmidrule(lr{8pt}){3-3} \cmidrule(lr{8pt}){4-4}  \cmidrule(lr{8pt}){5-5} \cmidrule(lr{8pt}){6-6} \cmidrule(lr{8pt}){7-7} \cmidrule(lr{8pt}){8-8} \cmidrule(lr{8pt}){9-9} \cmidrule(lr{8pt}){10-10} \cmidrule(lr{8pt}){11-11} \cmidrule(lr{8pt}){13-14}
    & CIDEr $\uparrow$
    & CIDEr $\uparrow$
    & CIDEr $\uparrow$
    & Val $\uparrow$
    & Val $\uparrow$
    & Val $\uparrow$
    & Val $\uparrow$
    & Val $\uparrow$
    & SPIDEr $\uparrow$
    & CIDEr $\uparrow$
    &
    & Rank $\downarrow$
    & Score $\uparrow$
    \\

\midrule

\moours{\limber}{(1)}{}{}
    & 122.85       
    & \ra{87.10}        
    & 51.85        
    & 60.19        
    & 45.93        
    & 17.96             
    & 33.38             
    & 34.13             
    & 38.94             
    & 46.03             
    & \ra{1h19}               
    & 8.0 (9)                       
    & 84.7 (9)                       
    \\
\moours{\limber}{(all)}{}{}
    & \ra{136.31}        
    & 83.74        
    & \ra{75.51}        
    & \ra{73.42}        
    & \rc{47.98}        
    & \ra{31.25}             
    & \rc{36.19}             
    & \ra{38.93}             
    & 40.12             
    & 46.87$^\blacksquare$             
    & 4h59               
    & \rb{3.2} (2)                       
    & \ra{97.4} (1)                       
    \\
\moours{\mapl}{}{}{}
    & 126.05        
    & 69.20        
    & 50.57        
    & 67.13        
    & 45.94        
    & 21.04             
    & \rb{36.21}             
    & \rc{37.02}             
    & 40.89             
    & 47.27             
    & 1h31               
    & 6.4 (7)                       
    & 86.9 (7)                       
    \\
\moours{\epalm}{}{}{}
    & 115.34        
    & 64.65        
    & 42.58        
    & 59.34        
    & 41.38        
    & 16.59             
    & 23.52             
    & 27.82             
    & 38.13             
    & 38.83             
    & \rb{1h20}               
    & 10.4 (10)                       
    & 73.1 (11)                       
    \\ 

\midrule
\suniLzero{}{}
    & 131.29        
    & \rb{87.05}        
    & \rb{73.67}        
    & \rc{70.11}        
    & \rb{48.25}        
    & \rc{22.97}             
    & \ra{37.69}             
    & \rb{38.45}             
    & \ra{43.37}             
    & \rc{49.88}             
    & \rc{1h25}               
    & \ra{2.5} (1)                       
    & \rb{95.7} (2)                       
    \\
\suniInner{}{}
    & 130.91        
    & 75.86        
    & \rd{65.22}        
    & 67.88        
    & 45.27        
    & \rb{23.70}           
    & \rd{35.98}           
    & \rd{36.36}           
    & \rd{41.20}             
    & 47.76             
    & 2h21               
    & 5.2 (5)                       
    & 90.7 (5)                       
    \\
\suniRavgpool{}{}
    & 131.77        
    & \rd{86.09}        
    & 61.18        
    & 64.84        
    & \ra{48.86}        
    & 19.14             
    & 35.17             
    & 35.41             
    & \rb{41.54}             
    & \rb{50.52}             
    & 1h50               
    & \rc{4.4} (3)                       
    & 90.4 (6)                       
    \\
\suniRlinear{}{}
    & \rc{133.01}        
    & 85.31        
    & \rc{69.76}        
    & 64.76        
    & \rd{47.66}        
    & 19.08             
    & 34.58             
    & 35.30             
    & 40.92             
    & \ra{51.60}             
    & 1h48               
    & 5.2 (5)                       
    & \rc{91.1} (3)                       
    \\
\suniRqsformer{}{}
    & \rd{131.92}        
    & 75.46        
    & 51.03        
    & 61.09        
    & 46.08        
    & 18.56             
    & 35.35             
    & 35.63             
    & 41.17             
    & 45.49             
    & 1h48               
    & 6.8 (8)                       
    & 85.6 (8)                       
    \\
\suniRrand{}{}
    & \rb{134.90}        
    & \rc{86.84}        
    & 58.15        
    & \rb{71.33}        
    & 47.60        
    & \rd{21.28}           
    & 35.00           
    & 34.74           
    & \rc{41.37}             
    & \rd{47.90$^\blacksquare$}             
    & 2h40               
    & \rc{4.4} (3)                       
    & 90.9 (4)                       
    \\
\sunicattn{}{}$^\blacklozenge$
    & 130.05        
    & 81.38        
    & --       
    & \rd{69.45}        
    & 41.73       
    & --          
    & --          
    & --          
    & --             
    & --             
    & 1h31               
    & 9.5$^\blacklozenge$ (10)                       
    & 36.9$^\blacklozenge$ (11)                       
    \\

\bottomrule
\end{tabular}
}

\label{tab:overall_results}
\end{table}

\paragraph{Better adaptation mechanism.}
Next, we explore seven additional  cross-modal interaction mechanisms, beyond the baseline ones, across a set of ten tasks. 
We also add  \limber\SP{(1)}, as it was shown to be a fast and efficient baseline~\cite{epalm}.
We report  results in \Cref{tab:overall_results}, where we also report the training time for AudioCaps as an illustration of the training cost.
To easily compare these methods across different tasks, we use two aggregate metrics.
(i) The average rank: for each task, we rank from 1 (best) to 11 (worst), and average the ranks across tasks.
(ii) The average score: we normalize the score for each task by the maximum score across the methods, and then average the normalized scores.

We use our results to conduct an analysis over the building blocks of the models.
First, using the same feature mapping, injecting tokens inside the LLM in the first-layer (\suniLzero{}{}, \limber\SP{(1)}) prevails over inner-layer injection (\epalm{}, \suniInner{}{}).  
We also found that cross-attention (\sunicattn{}{}) leads to unstable training in most low-data settings.
Extracting tokens from different encoder layers (\epalm{}, \suniInner{}{}, \sunicattn{}{}) makes sense with inner-layers injection techniques, but is not sufficient to improve over methods using only tokens from the last layer (\limber{} variants, \suniLzero{}{}, and  \sunibase{*,L0}{} variants).
Therefore, we now consider models with last-layer extraction and first-layer injection.
For the central mapping block, using a resampler (\sunibase{R-*,L0}{} and \sunibase{QP,*}{} variants) to reduce the number of tokens provides a trade-off between efficiency and performance, compared to injecting all tokens (\limber\SP{(all)}) or just one (\limber\SP{(1)} and \epalm{}). 
The \QPMapper{} used over all feature tokens (\suniLzero{}{} and \suniInner{}{}) provides the best trade-off, while  the local resamplers (\sunibase{R-*,L0}{}) that  preserve spatial feature structure lag behind or do not consistently achieve high scores.

Overall, \suniLzero{}{} and \limber\SP{(all)} achieve the best performance, reaching the best average rank and score, respectively.
In terms of training  speed,  however, \suniLzero{}{} is almost 4$\times$ more efficient, due to the small number of visual tokens injected in the LLM. 
Its training cost is similar to the most efficient approaches, \epalm{} and \limber\SP{(1)} that inject only one visual token, while significantly outperforming them.

\begin{figure}[h]
\centering
\begin{minipage}{.8\linewidth}

    \begin{minipage}{.33\linewidth} \centering
        \begin{imagebox} \centering
          \fbox{\includegraphics[height=\qualimgwidth,width=\qualimgwidth]{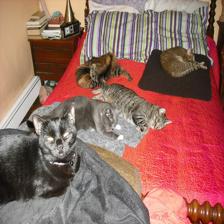}}
        \end{imagebox}
    \end{minipage}%
    \begin{minipage}{.33\linewidth}\centering
        \begin{imagebox} \centering
          \fbox{\includegraphics[height=\qualimgwidth,width=\qualimgwidth]{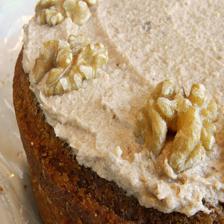}}
        \end{imagebox}
    \end{minipage}%
    \begin{minipage}{.33\linewidth}\centering
        \begin{imagebox} \centering
          \fbox{\includegraphics[height=\qualimgwidth,width=\qualimgwidth]{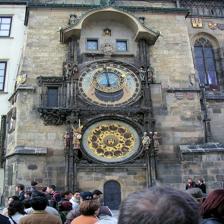}}
        \end{imagebox}
    \end{minipage}
    
    \qualtext{A group of cats laying on top of a bed.}%
    \qualtext{A close up of a cake with nuts on it}%
    \qualtext{A group of people standing in front of a large clock.}%

        \vspace{5mm}

    \begin{minipage}{.33\linewidth} \centering
        \begin{imagebox} \centering
          \fbox{\includegraphics[height=\qualimgwidth,width=\qualimgwidth]{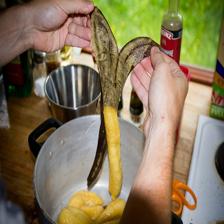}}
          
          \questiontext{What is the substance in the bowl?}
        \end{imagebox}
    \end{minipage}%
    \begin{minipage}{.33\linewidth}\centering
        \begin{imagebox} \centering
          \fbox{\includegraphics[height=\qualimgwidth,width=\qualimgwidth]{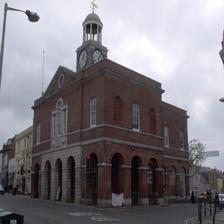}}
          
          \questiontext{What color is the building?}
        \end{imagebox}
    \end{minipage}%
    \begin{minipage}{.33\linewidth}\centering
        \begin{imagebox} \centering
          \fbox{\includegraphics[height=\qualimgwidth,width=\qualimgwidth]{figures/qualitative/vqav2/115.jpg}}
          
          \questiontext{How many rolls are on the plate?}
        \end{imagebox}
    \end{minipage}
    
    \qualtext{bananas}%
    \qualtext{brown}%
    \qualtext{2}%

    \vspace{5mm}
    
    \begin{minipage}{.33\linewidth} \centering
        \begin{imagebox} \centering
          \includegraphics[height=16pt,width=16pt]{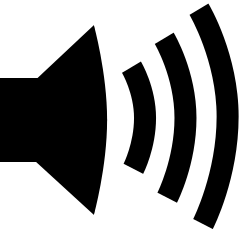}
          
          \questiontext{A sewing machine running briefly}
        \end{imagebox}
    \end{minipage}%
    \begin{minipage}{.33\linewidth}\centering
        \begin{imagebox} \centering
          \includegraphics[height=16pt,width=16pt]{figures/qualitative/audio.png}
          
          \questiontext{A vehicle engine is revving up}
        \end{imagebox}
    \end{minipage}%
    \begin{minipage}{.33\linewidth}\centering
        \begin{imagebox} \centering
          \includegraphics[height=16pt,width=16pt]{figures/qualitative/audio.png}
          
          \questiontext{A woman is speaking while food is frying and sizzling}
        \end{imagebox}
    \end{minipage}
    
    \qualtext{A sewing machine is being used}%
    \qualtext{A vehicle engine is idling and then accelerates}%
    \qualtext{A woman is speaking and frying food}%
\end{minipage}

\caption{
Qualitative samples of DePALM on  COCO (first row), VQAv2 (second row) and AudioCaps (third row), when finetuned on 100\% of the data. Input is in gray, and output is in green. For  AudioCaps, a ground-truth caption of the input is shown.
}
\label{fig:qualitative_expes}
\end{figure}

\paragraph{Qualitative results.}
We give some qualitative results of our model for multiple multimodal tasks in \Cref{fig:qualitative_expes}.
We can notice that the models adapt to answer in the  style corresponding to the dataset, and has notions of real-world objects, being able to identify colors, animals and objects.

\begin{table}
\small
\centering  
    \caption{
        Comparison of different visual backbones, with a fixed LLM (left) and with different LLMs (right). We show the CIDEr score on COCO, the validation accuracy on OKVQA, and the SPICE score on AudioCaps.
        For reference, we add the ImageNet~\cite{deng2009imagenet} Top1 score of each visual backbone, and ARC for each LLM, that measures textual question answering  capabilities. The results are averaged over 3 runs.}
            
    \resizebox{0.49\textwidth}{!}{%

        \label{tab:ab_backbones_visual}
        \begin{tabular}{lcccc}
        \toprule        
        Visual backbone
            & \makecell{COCO\\\suniLzero{}{}}
            & \makecell{COCO\\\limber\SP{(1)}}
            & \makecell{OKVQA\\\suniLzero{}{}}
            & \makecell{ImageNet\\Top1}
            \\
        \midrule
        
        DINOv2-S~\cite{dinov2}
            & 118.26    
            & 100.63    
            & 33.64    
            & 81.1\%    
            \\
        
        DINOv2-B~\cite{dinov2}
            & 125.42    
            & 106.12    
            & 34.82    
            & 84.5\%    
            \\
        
        DINOv2-L~\cite{dinov2}
            & 126.95    
            & 107.17    
            & 31.81    
            & \rb{86.3}\%    
            \\
        
        DINOv2-G~\cite{dinov2}
            & \rc{127.49}    
            & 110.58    
            & 35.52    
            & \ra{86.5}\%    
            \\
        
        ViT-L~\cite{steiner2021augreg}
            & 118.59    
            & 106.49    
            & 36.11    
            & \rc{85.6}\%    
            \\
        
        CLIP-ViT-B~\cite{clip}
            & 121.93    
            & \rc{111.88}    
            & \rc{36.68}    
            & 68.6\%    
            \\
        
        CLIP-ViT-L~\cite{clip}
            & \rb{128.69}    
            & \rb{116.80}    
            & \ra{37.27}    
            & 75.3\%    
            \\
               
        EVA-CLIP-L~\cite{evaclip}
            & \ra{130.66}    
            & \ra{123.20}    
            & \rb{37.13}    
            & 79.8\%    
            \\
        
        \bottomrule
        \end{tabular}
    }
     \hfill
    \resizebox{0.49\textwidth}{!}{%
    
        \label{tab:ab_backbones_llm}
        \begin{tabular}{lcccc}
        \toprule        
        \makecell{LLM\\backbone}
            & \makecell{COCO\\\suniLzero{}{}}
            & \makecell{COCO\\\limber\SP{(1)}}
            & \makecell{AudioCaps\\\suniLzero{}{}}
            & ARC
            \\
        \midrule
        
        OPT-125M~\cite{opt}
            & 126.88    
            & 102.45    
            & 41.82    
            & 22.87    
            \\
        
        OPT-1.3B~\cite{opt}
            & \rc{129.41}    
            & {112.43}    
            & \rc{42.77}    
            & 29.52    
            \\
        
        OPT-2.7B~\cite{opt}
            & {125.75}    
            & \rc{115.81}    
            & \rb{43.35}    
            & 33.96    
            \\
        
        OPT-6.7B~\cite{opt}
            & \ra{131.64}    
            & \rb{117.51}    
            & \ra{43.83}    
            & \rc{39.16}    
            \\
        
        \llama{}-7B~\cite{llama}
            & \rb{130.73}    
            & \ra{123.12}    
            & 42.48    
            & \rb{51.02}    
            \\
            
        Vicuna-7B~\cite{vicuna}
            & 125.66    
            & 111.53    
            & 21.79    
            & \ra{53.24}    
            \\  
        \bottomrule
        \end{tabular}
    }

\end{table}

\subsection{Analysis and ablation study}

\paragraph{Text-aligned perceptual features adapt better to LLMs.}
We investigate the influence of the perceptual backbones on the overall performance. 
In \Cref{tab:ab_backbones_visual} (left)  
we compare different visual encoders with varying sizes and different training paradigms on different image captioning and visual question answering datasets. 
For the same model family, see DINOv2 and CLIP-ViT, the bigger size the better the performance. 
Self-supervised encoders (DINOv2) performed better than supervised ones (ViT) for image captioning, but the reverse was true for OKVQA.
Finally, vision-language pre-training of the encoders (CLIP) performs best across all tested settings. 
This reveals that, using existing text-aligned perceptual encoders, makes the cross-modal interaction between the encoder and LLMs more effective.
Overall, models with better feature quality (higher ImageNet score) increase our results, with a large boost when there is a pre-existing alignment with text.

\paragraph{Better LLMs are not always  better for multimodality.}
Next, in \Cref{tab:ab_backbones_llm} (right), we compare LLMs with different models and pretraining data sizes, and consider the impact on image and audio captioning results.  
We find a clear positive correlation between the LLM size  and the score for the OPT models, similar to the  ARC metric~\cite{arc_metric} which measures  textual question-answering capabilities of the LLMs.
However, when comparing LLMs with similar model sizes in the 7B range, we do not see a clear improvement when using LLMs pretrained on more data (LLaMA), nor when fine-tuning on language instructions (Vicuna), contrary to observations for the ARC metric. 

\begin{figure}[t]
   \centering
       \includegraphics[width=.46\linewidth]{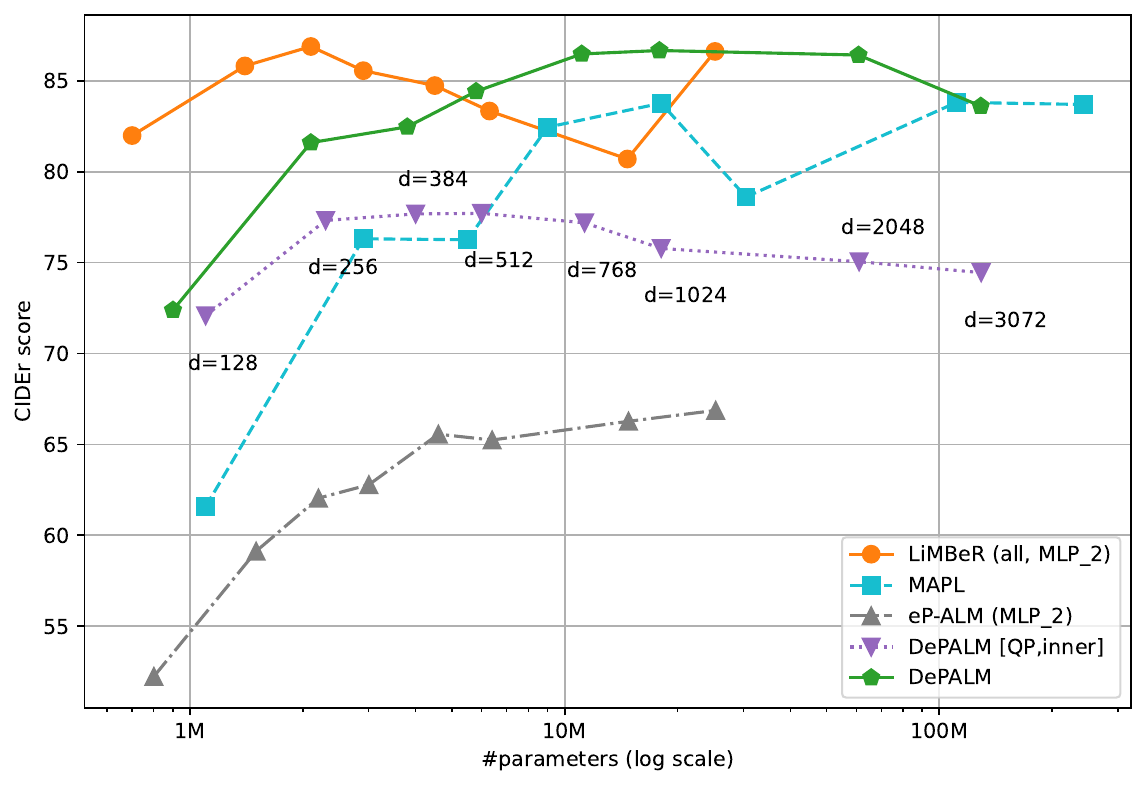}
    \hfill
           \includegraphics[width=.49\linewidth]{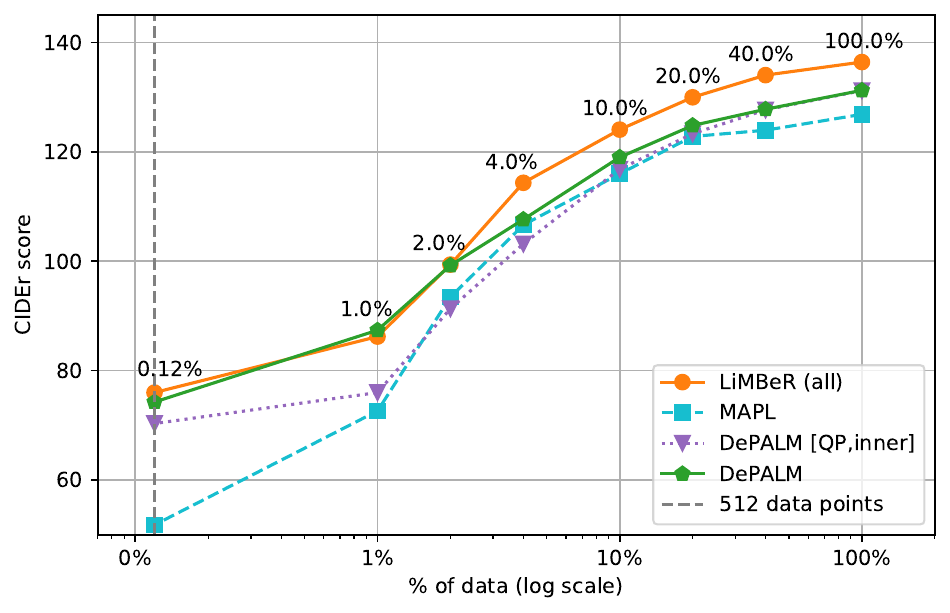}
   \caption{Captioning performance on COCO (averaged across three runs) as a function of the number of parameters when using 1\% of the training data  (left),  and     as a function of the training set size (right).
   We control the number of parameters  by setting  the internal feature dimension $d$. For LiMBeR and eP-ALM, we replace the single linear projection with a two-layer projection 
   (MLP\_2) with varying number of hidden units.}
          \label{fig:cider_for_nb_params}
       \label{fig:cider_for_training_size}
\end{figure}

\paragraph{Parameter and data efficiency.}
We consider COCO captioning performance as a function of the number of trainable parameters in \Cref{fig:cider_for_nb_params} (left), for a set of diverse methods: we include the two best architectures (\limber\SP{(all)} and \suniLzero{}{}) and add a set of diverse models using different injection or mapping mechanisms (\suniInner{}{}, \epalm{}, \mapl{}) to compare to diverse behaviors.
We focus on the low-data training regime by using only 1\% of the training set.
To vary the number of trainable parameters, we do the following: for \mapl{} and \sunibase{}{} we change the hidden dimension of the \QPMapper{}, for \limber\SP{(all)} and \epalm{},we replace the linear feature projection by a two-layer projection (MPL\_2) with a bottleneck of variable dimension. 

First, we observe that for most of the considered parameter range,  \limber\SP{(all)} and \suniLzero{}{} yield the best performance, coherent with earlier experiments.
Second, we do not observe strong overfitting for any of the methods, suggesting that in this small data regime the type of interfacing mechanism is more important for  performance than the number of parameters. 

We  investigate  data efficiency by varying the training data size from 0.12\% to 100\% in \Cref{fig:cider_for_training_size} (right). 
All methods scale similarly well with the  number of training examples, with \limber\SP{(all)} and \suniLzero{}{} yielding optimal performance across all data sizes.
We find that training only on 10\% of data achieves roughly 90\% of the final performance, validating the data-efficiency of these methods. 

\begin{table}[t]
\centering	
\setlength\tabcolsep{4pt}
\caption{
    Comparison with state-of-the-art LLM augmentation methods. The DePALM results are averaged over 3 runs.
    We highlight the \ra{best} results for each category in underlined bold. 
    $\dagger$: use standard split instead of the Karpathy one.
    Note that only the results in the last group of parameter efficient methods are directly comparable to ours.
     }
\resizebox{0.8\linewidth}{!}{%
\begin{tabular}{lccccc}
\toprule	 	
\multirow{2}{*}{Method}
    & COCO
    & COCO (1\%)
    & VQAv2
    & OKVQA
    & MSRVTT
    \\
\cmidrule(lr{8pt}){2-2} \cmidrule(lr{8pt}){3-3} \cmidrule(lr{8pt}){4-4}  \cmidrule(lr{8pt}){5-5} \cmidrule(lr{8pt}){6-6}
    & CIDEr
    & CIDEr
    & Val
    & Val
    & CIDEr
    \\
\midrule
\rowcolor{llgray} \multicolumn{6}{c}{Large-scale methods in few-shot mode}\\
\midrule

\rowcolor{llgray} \flamingo{}~\cite{flamingo} (32-shot)
    & 113.8        
    & --        
    & \ra{67.6}        
    & \ra{57.8}        
    & --        
    \\ 
\rowcolor{llgray}\bliptwo{}~\cite{blip2} (0-shot)
    & \ra{121.6}        
    & --        
    & --        
    & 45.9        
    & --        
    \\

\midrule
\rowcolor{lllgray}\multicolumn{6}{c}{Large-scale methods finetuned on target task}\\
\midrule

\rowcolor{lllgray}\flamingo{}~\cite{flamingo} 
    & 138.1$^\dagger$        
    & --        
    & 82.1$^\dagger$        
    & --        
    & --        
    \\

\rowcolor{lllgray}\bliptwo{}~\cite{blip2} 
    & 145.8        
    & --        
    & \ra{82.30}$^\dagger$         
    & --        
    & --        
    \\

\rowcolor{lllgray}UnIVAL~\cite{unival} 
    & 128        
    & --        
    & 73.24        
    & 45.7        
    & \ra{56.3}        
    \\

\rowcolor{lllgray}NExT-GPT~\cite{Wu2023NExTGPTAM} 
    & \ra{156.7}        
    & --        
    & --        
    & --        
    & --        
    \\
\rowcolor{lllgray}IDEFICS 80B Instruct~\cite{laurenccon2024obelics} (32-shot)
    & 123.2        
    & --        
    & 68.8        
    & \ra{59.5}         
    & --        
    \\
\rowcolor{lllgray}Qwen-VL (7B)~\cite{bai2023qwenvl}
    & --        
    & --        
    & 79.5        
    & 58.6         
    & --        
    \\

\midrule
\multicolumn{6}{c}{Parameter-efficient methods for LLM augmentation}\\
\midrule

\mapl{}~\cite{epalm}
    & 125.2        
    & 65.9        
    & 43.5        
    & 31.6        
    & --        
    \\

\vladapter{}~\cite{vl-adapter}
    & 116        
    & --        
    & 65.9        
    & --        
    & --        
    \\

\epalm{}~\cite{epalm}
    & 111.6        
    & --        
    & 54.9        
    & --        
    & 48.8        
    \\

\suniLzero{}{} (ours)
    & \ra{131.3}        
    & \ra{87.1}        
    & \ra{70.1}        
    & \ra{37.7}             
    & \ra{49.9}             
    \\

\bottomrule
\end{tabular}
}

\label{tab:sota}
\end{table}

\paragraph{Comparison with the state-of-the-art.} 
In \Cref{tab:sota} we compare our results with state-of-the-art approaches, including large-scale ones. 
We compare only with models with at least one top score.
For this comparison, we use \suniLzero{}{} with the backbones listed in \Cref{tab:datasets}.
Our approach outperforms all parameter-efficient approaches (bottom part of the table) such as \epalm{} and \mapl{}. 
We significantly reduce the performance gap of parameter-efficient approaches \wrt to large-scale models that are fine-tuned to the target task (middle part of the table). We also compare our approach to generalist models  that do not require finetuning (top part), showing that we compete and sometimes outperform them.  While, these models are not directly comparable to ours, the results show that  finetuning can significantly boost performance,   and \suniLzero{}{} emerges as a promising and efficient approach, that does not require large-scale pretraining.
\section{Discussion and conclusion}

\paragraph{Small \vs large-scale setups.} This work focuses on adapting LLMs for multimodal tasks with focus on efficiency along three main axes:  (a)  training set size, (b) number trainable parameters, and (c) amount of  compute. 
This allows to obtain LLM-based solutions significantly faster and  more affordably. 
Importantly, it streamlines  the adoption of  stronger LLMs and perceptual foundation models that are continuously released. 

A different approach is to go large scale  along these three  axes, with the objective to obtain good performance across many datasets~\cite{liu2023improvedllava15,bai2023qwenvl,liu2024visualinstructblip,blip2,flamingo,chen2023pali}. 
This usually requires conducting pretraining, followed by instruction tuning, and even single-task finetuning when targeting a particular dataset. While this approach is more generalist, 
it requires enormous resources in terms of data and compute. Nonetheless, we believe that both setups are worth pursuing and are complementary, paving the way for very effective multimodal models, spanning a wide range of setups.       

\paragraph{Limitations.} While this work achieves large improvements compared to previous efficient approaches, there is still room for improvement, especially regarding harder tasks such OK-VQA or those requiring  reasoning \cite{lu2022learnscienceqa}. We believe the proposed framework, will be a good ground to develop more effective approaches in the future. Besides, this work focuses mainly on performance and efficiency. However, there are other axes that should be considered before deployment. In particular, safety issues, such hallucinations \cite{shukor2024beyond,liu2023aligning}, abstention \cite{Dancette_2023_CVPR}, harmfulness or the broader objective of aligning these models to human preferences \cite{sun2023aligningrlhffact}. 

\paragraph{Conclusion.}
We presented a systematic comparative study of mechanisms to interface perceptual backbones ---for image, video and audio data--- with large language models to address tasks such as captioning and  question answering. 
We focus on parameter efficient approaches, which leave the LLM and feature backbone unchanged, and can be trained on limited training sets.
We conducted extensive experiments on different  datasets and tasks in which we evaluated both existing and new mechanisms, considered different choices for the perceptual backbones and language models, and tune hyperparameters for all methods in a fair manner. 
We find improved results as compared to previously reported ones, even when using the same existing interfacing mechanisms.
In general, our study shows that most of the improvement, is coming from better perceptual encoders, especially text-aligned ones, in contrast to using more powerful LLMs. We also find that simple design choices works best, such as passing all perceptual tokens at the input to the LLM, or using transformer-based token pooling mechanisms for efficiency.
Moreover, we find that our proposed \suniLzero{}{} mechanism ---which compresses tokens from the perceptual backbone to a few ``summary tokens'' to inject in the LLMs--- yields on par or better results than existing approaches, while being 4$\times$ faster to train than the second best method. 

\bibliographystyle{splncs04}
\bibliography{main,jjv}

\appendix

\setcounter{figure}{0} 
\renewcommand\thefigure{S\arabic{figure}} 
\setcounter{table}{0}
\renewcommand{\thetable}{S\arabic{table}}

\section{Assets and licensing information}
\label{app:licensing}

In \Cref{tab:links}, we list the datasets and pre-trained models we use for our experiments. We provide the links to the to repositories and their licenses.

\begin{table*}[b]

\centering	
\setlength\tabcolsep{4pt}
{\scriptsize
\begin{tabular}{lcc}
\toprule
Name & Link & license \\
\midrule
COCO~\cite{coco} & \url{https://cocodataset.org} & CC BY 4.0 \\
TextCaps~\cite{textcaps} & \url{https://textvqa.org/textcaps/} & CC BY 4.0 \\
VQAv2~\cite{vqav2} & \url{https://visualqa.org/} & CC BY 4.0 \\
TextVQA~\cite{textvqa} & \url{https://textvqa.org/} & CC BY 4.0 \\
OKVQA~\cite{okvqa} & \url{https://okvqa.allenai.org/} & Unknown \\
AOKVQA~\cite{aokvqa} & \url{https://allenai.org/project/a-okvqa} & Unknown \\
AudioSet~\cite{audioset} & \url{https://research.google.com/audioset/} & CC BY 4.0 \\
AudioCaps~\cite{audiocaps} & \url{https://audiocaps.github.io/} & MIT \\
MSRVTT~\cite{msrvtt} & \makecell{
\href{https://www.microsoft.com/en-us/research/publication/msr-vtt-a-large-video-description-dataset-for-bridging-video-and-language/}{Microsoft website}
} & Unknown \\
\midrule
CLIP~\cite{clip} & \url{https://github.com/openai/CLIP} & Unknown \\
EVA-CLIP~\cite{evaclip} & \url{https://github.com/baaivision/EVA/tree/master/EVA-CLIP} & MIT \\
DINOv2~\cite{dinov2} & \url{https://github.com/facebookresearch/dinov2} & Apache License 2.0 \\
ViT-L~\cite{steiner2021augreg} & \url{https://github.com/huggingface/pytorch-image-models} & Apache License 2.0 \\
TimeSformer~\cite{timesformer} & \url{https://github.com/facebookresearch/TimeSformer} & CC BY 4.0 \\
OPT~\cite{opt} & \url{https://github.com/facebookresearch/metaseq} & MIT \\
LLaMA~\cite{llama} & \url{https://github.com/facebookresearch/llama} & \href{https://github.com/facebookresearch/llama/blob/main/LICENSE}{llama license} \\
Llama2~\cite{llama2} & \url{https://ai.meta.com/llama/} & \href{https://github.com/facebookresearch/llama/blob/main/LICENSE}{llama license} \\
Vicuna~\cite{vicuna} & \url{https://lmsys.org/blog/2023-03-30-vicuna/} & \href{https://github.com/facebookresearch/llama/blob/main/LICENSE}{llama license} \\
 bottomrule
\end{tabular}
}
\caption{Links to the assets used in the paper, and their respective licenses.}
\label{tab:links}
\end{table*}

\section{Building blocks of our  framework}
\label{app:Blocks}
In this section, we provide more details about the different blocks we use to implement existing baseline models, as well as our DePALM models. We suppose a feature extractor model with tokens of dimension $d_\text{feats}$, and a LLM with tokens of dimension $d_\text{LLM}$

\subsection{Feature extraction} 

The design of feature extraction is based on two decisions: the number $n_\text{fl}$ of feature levels, and whether we keep all tokens, or only the CLS token.
We take the output of the last $n_\text{fl}$ transformer layers of the feature extractor (the image, video or audio backbone).
It gives us an output of dimension $(n_\text{fl},n_\text{tk}+1,d_\text{feats})$, where $n_\text{tk}$ is the number of patch tokens, and $d_\text{feats}$ the embedding dimension of the feature model (perceptual encoders). When only extracting the CLS token, the output dimension becomes $(n_\text{fl},1,d_\text{feats})$ (the patch tokens are removed).

A special case is added for the MAViL model, where the CLS token is replaced by the mean of all patch tokens of the same level, but only when we do not keep the patch tokens.

\subsection{Feature injection}

\paragraph{First-layer token injection.} Here $n_\text{fl}=1$. The feature tokens are prepended to the sequence of textual token embeddings (including ``BOS'' token for OPT). They are propagated through the LLM, and removed from its final output. We use a causal attention mask, where each token can only attend to previous ones, including in-between inserted perceptual tokens.

\paragraph{Inner-layers token injection.} Here we only require $n_\text{fl}\geqslant 1$. This method is additionally parametrized by a number of LLM layers $n_\text{LLM}$ where we inject tokens and a number of left-out layers at the end $n_\text{left}$. Then, if we note $L_\text{LMM}$ the total number of LLM layers, for each layer $i\in \{L_\text{LMM} - n_\text{LLM} - n_\text{left}, \dots, L_\text{LMM} - n_\text{left}-1\}$, we inject feature tokens extracted from level $l_i=\lfloor \frac{(i-L_\text{LMM}) * n_\text{fl}}{n_\text{LLM}} \rfloor$. For injection, we follow the same procedure as for first-layer token injection, where the feature tokens are prepended to the input sequence of the layer $i$. Additionally, we remove them from the output sequence of this layer.

\subsection{Feature mapping}

\paragraph{\QPMapper.} This mapping block is parametrized by the number of layers $L_\textrm{QP}$ and the number of query tokens $n_\textrm{Q}$. It takes as input a sequence of tokens of dimension $d_\textrm{embed}$.
These tokens are concatenated to $n_\textrm{Q}$ query tokens, which are learnable parameters. The resulting sequence is then passed through a stack of $L_\textrm{QP}$ standard transformer encoder layers. We use a dropout of $0.1$, embedding dimension of $d_\text{embed}$, the GELU activation and 8 attention heads. Only the last $n_\textrm{Q}$ output tokens, corresponding to the query tokens, are considered as output.

\paragraph{Block-based token resamplers.} Some of the resamplers use a common framework, based on local blocks of patches. This framework is parametrized by an embedding dimension $d_\textrm{emb}$, and a pooling function. The extracted feature tokens are first projected from dimension $d_{\textrm{feats}}$ to $d_\textrm{embed}$ using a linear layer. The patches tokens are then arranged on a 1D, 2D or 3D grid, depending on the modality, and grouped into blocks of dimension $4$ (1D case) or $2\times 2$ (2D case). Each block is pooled using the pooling function, resulting in a single token. This is similar to using a 1D or 2D pooling operation on the grid. The tokens are then rearranged as a sequence again, to which the CLS token is prepended, before being normalized using the RMSNorm, and finally projected with a linear layer from $d_\textrm{embed}$ to $d_\textrm{LLM}$.

\subsection{Fine-tuning mechanism}

\paragraph{Prompt-tuning.} is a parameter-efficient fine-tuning mechanism, parametrized by $n_\textrm{pt}$, the number of learned tokens. When used, $n_\textrm{pt}$ constant embedding vectors of dimension $d_\textrm{LLM}$ are learned and prepended before the textual tokens (and after the perceptual tokens) at the beginning of LLMs. We also use a causal padding attention mask with these tokens.

\section{Implementation details}
\label{app:implementation}

\subsection{Reproducing existing baseline}
\label{app:Baselines}

\paragraph{\limber\SP{(all)}:} we use the feature extraction with $n_\text{fl}=1$ level, project all the feature tokens from dimension $d_{\text{feats}}$ to $d_\text{LLM}$ using a linear layer, and use first-layer token injection.

\paragraph{\limber\SP{(1)}:} we use the same mechanism as for \limber\SP{(all)}, but only keep the single CLS token.

\paragraph{\mapl{}:} we use the feature extraction with $n_\text{fl}=1$ level, and a feature mapping block consisting of a linear projection from dimension $d_{\text{feats}}$ to $d_\text{embed}=256$, followed by a \QPMapper{} using $L_\textrm{QP}=4$ layers and $n_\textrm{Q}=32$ query tokens, and then a linear projection from $d_\text{embed}$ to the LLM inner dimension $d_\text{LLM}$. We then insert tokens with the first-layer injection mechanism.

\paragraph{\epalm{}:}
we extract $n_\text{fl}=6$ levels of feature tokens, and only keep the CLS token from each level. We project each one from  dimension $d_{\text{feats}}$ to $d_\text{LLM}$ with the same linear layer.
Mapped tokens are inserted into $n_\text{LLM}=12$ inner layers, leaving out the last one ($n_\text{left}=1$).

\subsection{DePALM variants}
\label{app:DePALM}

Our DePALM and  DePALM\SP{*,L0} methods 
use the following blocks:
\begin{itemize}
    \item Feature extraction from $n_\text{fl}=1$ level.
    \item First-layer injection of token, after the mapping block.
    \item Prompt-tuning with $n_\text{pt}=1$.
\end{itemize}

\paragraph{DePALM:} we use a feature mapping block consisting of a linear projection from dimension $d_{\textrm{feats}}$ to $d_\text{embed}=1024$, followed by a \QPMapper{} using 
$L_\textrm{QP}=2$ layers and $n_\textrm{Q}=32$ query tokens, and a linear projection from $d_\textrm{embed}$ to the LLM inner dimension $d_\textrm{LLM}$.

\paragraph{DePALM\SP{R-rand,L0}:} we sample a subset of the tokens, using the following procedure. If the model outputs $n_\text{tk}$ patch tokens, we keep the CLS token, and $\lfloor f n_\text{tk} \rfloor$ uniformly sampled patch tokens, with $f$ the proportion of tokens we keep:
\begin{itemize}
    \item During training, with a probability of $\frac{1}{10}$, we set $f=f_\text{max}$. Otherwise, we sample $f'\sim \mathcal{N}(f_\text{mean},f_\text{std})$, and set $f=\min(\max(f,f_\text{min}),f_\text{max})$ to restrict the values to the interval $[f_\text{min};f_\text{max}]$. 
    \item During inference, we set $f=f_\text{max}$.
    \item We use $f_\text{min}=\frac{1}{16}$, $f_\text{max}=\frac{1}{2}$, $f_\text{mean}=\frac{1}{4}$ and $f_\text{std}=0.2$.
\end{itemize}
We project the resulting tokens from dimension $d_{\text{feats}}$ to $d_\text{LLM}$, normalize them using the RMSNorm, and project them again with a linear layer ($d_\text{LLM}$ to $d_\text{LLM}$) before injection inside the LLM.

\paragraph{DePALM\SP{R-linear,L0}:} we use a \textit{block-based resampler} with $d_\text{emb}=d_\text{LLM}$. The pooling function is a linear projection from dimension $4\times d_\text{emb}$ to $d_\text{emb}$, taking as input the concatenation of all tokens from the same block.

\paragraph{DePALM\SP{R-\QPMapper{},L0}:} we use a \textit{block-based resampler} with $d_\text{emb}=768$. The pooling function is a \QPMapper{} taking as input the 4 tokens of a single block, using $L_\textrm{QP}=4$ layers and a single query token ($n_\textrm{Q}=1$).

\paragraph{DePALM\SP{R-avgpool,L0}:} we use a \textit{block-based resampler} with $d_\text{emb}=d_\text{LLM}$. The pooling function returns the mean of the tokens from the same block.

\paragraph{DePALM\SP{QP,inner}:} we extract $n_\text{fl}=4$ levels of feature tokens, and use the same feature mapping as \suniLzero{}{}. In particular, the mapping block is shared across all token levels, meaning that each sequence of tokens from each level goes through the same projection, with the same weights, but as separate batch elements.
Mapped tokens are inserted into $n_\text{LLM}=12$ inner layers, with $n_\text{left}=3$.
We also use prompt-tuning with $n_\textrm{pt}=16$.

\paragraph{DePALM\SP{c-attn}:} we extract $n_\text{fl}=4$ levels of feature tokens, and use a single cross-attention block (detailed below) that will do both feature mapping and injection.
Injection takes place into the last $n_\text{LLM}=12$ inner layers, leaving out the $n_\text{left}=3$ last ones, similarly to the inner-layer injection mechanism, but using cross-attention instead of concatenation.

\begin{figure}[t]
   \centering
   \includegraphics[width=0.7\linewidth]{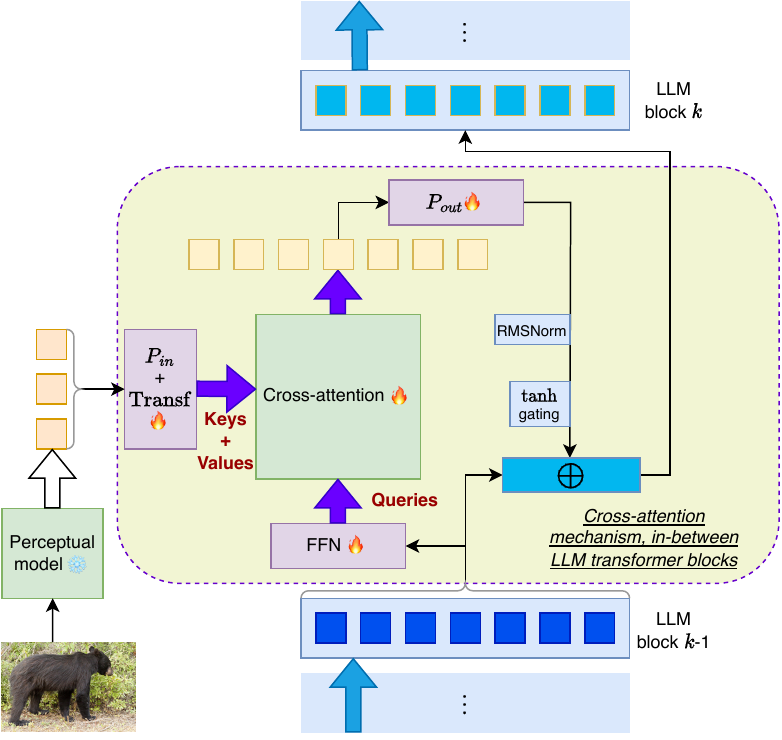}
   \caption{Overview of cross-attention mapping and injection  in DePALM\SP{c-attn}.}
   \label{fig:cross-attention}
\end{figure}

For the cross-attention block, see \Cref{fig:cross-attention}, we first project the tokens from dimension $d_{\text{feats}}$ to $d_\text{embed}=1024$ using the linear projection $P_\textrm{in}$.
They are then passed through a single transformer layer $\mathrm{Transf}$ (8 heads, dropout of 0.1),
acting as a minimal resampler network. We also project the input textual tokens to the dimension $d_\text{embed}=1024$, using a two-layers MLP (the FFN block, with a hidden dimension of $d_\text{embed}$ and the GELU activation function), and normalize them using RMSNorm. We then use the perceptual tokens as keys and values, and the text ones as queries, in a cross-attention layer. We project the resulting tokens back to dimension $d_\text{LLM}$ using a linear projection, normalize them with RMSNorm, and add them to input textual tokens, using a tanh-gated residual connection. This residual connection takes the form $x^{(k-1)},x^{(k)}_r,h^{(k)} \mapsto x^{(k-1)} + \tanh (h^{(k)}) \times x^{(k)}_r$ where $x^{(k-1)}$ are the original textual tokens, $x^{(k)}_r$ the output of our cross-attention block, and $h^{(k)}$ a single learned float, initialized at $0$.

\subsection{Training}
\label{app:training}

\paragraph{General training.} We train each model on a single node using eight V100-32G GPUs, using the following method:
\begin{itemize}
    \item Optimizer: we take AdamW, with a weight decay of 0.1 and a default learning rate of $\alpha_\text{max}=8\cdot 10^{-4}$, that we further adapt for each experiment.
    \item Gradient clipping: we use a clipping value of $g_\text{clip}=0.8$, that we further adapt for each experiment.
    \item Learning rate scheduler: we set a minimum learning rate of $\alpha_\text{min}=\alpha_\text{max}\cdot 10^{-4}$, with a cosine scheduler. During the first 20\% of all iteration steps, we linearly warmup the effective learning rate from $\alpha_\text{min}$ to $\alpha_\text{max}$, then use the cosine scheduler to decrease it from $\alpha_\text{max}$ to $\alpha_\text{min}$.
    \item Batch size: we use a batch size of 16 on each GPU, for an effective batch size of 128. On experiments where memory is an issue, we use gradient accumulation while to train on the same batch size.
    \item Epochs: we use a base number of 8 epochs, and increase it on small datasets: we use 12 epochs on AudioCaps, 20 on TextCaps, OKVQA, AOKVQA and TextVQA, and 30 epochs on the two 1\% settings.
    \item Loss: we compute the loss only on the generate text. Additionally, we use a label smoothing value of $2\cdot 10^{-3}$ in the cross-entropy loss of the LLM.
    \item Float precision: we load pre-trained models in float16, and train new weights in float32, using mixed-precision.
    \item Duplicate inputs: we group together every training sample with the same perceptual and text input, but different outputs. During training, we select the target output for the loss randomly.
\end{itemize}

\paragraph{Grid search.}
As each setting has a different training dynamic, which can be very sensitive to the learning rate, we use a grid search over a few values that we experimentally found to be working efficiently. 
We start by sweeping over the learning rate values $\alpha_\text{max}\in \{1\cdot 10^{-3}, 8\cdot 10^{-4},  4\cdot 10^{-4}  \}$, and set the gradient clipping  parameter as $g_\text{clip}=0.8$. 
On each experiment where the score significantly increase or decrease between the three runs, or where the results are noticeably lower than other methods or previous experiments, we further experiment with $g_\text{clip}=0.8$ for the same sweep over learning rates. 
We perform this sweep  for each method-dataset pair, including baselines reproductions.

We also note that we used grid search on each parameter of our models (number of layers, etc.) on the COCO dataset, to confirm that they are at least a local optimum.

\paragraph{Data augmentation.}
We use random augmentation during training of each model. Each time a training sample is used during the training loop, a random modification is applied, without increasing the dataset size.
\begin{itemize}
    \item Images: the normalized image is resized to $128\times 128$ with a random scale in $[0.5, 1]$ and a random aspect ratio in $[\frac34,\frac43]$, and horizontally flipped with probability $0.5$. Data is then augmented with the modified RandAugment procedure~\cite{ALBEF}.
    \item Audio: we augment the dataset using frequency masking with maximum length of 24, and time masking with maximum length of 96, after normalizing the audio.
    \item Video: we normalize the videos, and use the same procedure as with images, with the same random scaling and flipping as with images, followed by the default RandAugment procedure implemented in pytorch.
\end{itemize}

\paragraph{Special cases:} due to memory limitations, when training \limber\SP{(1)} and \suniRrand{}{} on MSRVTT, we average the patch tokens along the time dimension to reduce their number. This yields a representation with the dimension of the embedding of a single frame.

\section{Experimental results}
\label{app:results}

\begin{table*}
\small
\centering	
\setlength\tabcolsep{4pt}
\resizebox{\linewidth}{!}{%
\begin{tabular}{lcc<{\hspace{8pt}}cc<{\hspace{8pt}}cc<{\hspace{8pt}}cc<{\hspace{8pt}}cc<{\hspace{8pt}}c<{\hspace{8pt}}c<{\hspace{8pt}}c<{\hspace{8pt}}}
\toprule	 	
\multirow{2}{*}{Method}
    & \multicolumn{2}{c<{\hspace{8pt}}}{COCO}
    & \multicolumn{2}{c<{\hspace{8pt}}}{COCO (1\% data)}
    & \multicolumn{2}{c<{\hspace{8pt}}}{TextCaps}
    & \multicolumn{2}{c<{\hspace{8pt}}}{VQAv2}
    & \multicolumn{2}{c<{\hspace{8pt}}}{VQAv2 (1\% data)}
    & TextVQA
    & OKVQA
    & AOKVQA
    \\
\cmidrule(lr{8pt}){2-3} \cmidrule(lr{8pt}){4-5} \cmidrule(lr{8pt}){6-7}  \cmidrule(lr{8pt}){8-9} \cmidrule(lr{8pt}){10-11} \cmidrule(lr{8pt}){12-12}  \cmidrule(lr{8pt}){13-13}  \cmidrule(lr{8pt}){14-14}
    & B@4 & CIDEr
    & B@4 & CIDEr
    & B@4 & CIDEr
    & Val & Test
    & Val & Test
    & Val
    & Val
    & Val
    \\

\midrule

\moours{\limber}{(1)}{}{} & 35.85±0.32 & 122.85±0.69 & 25.05±0.31 & 87.10±1.36 & 17.28±0.08 & 51.85±0.71 & 60.19±0.36 & 59.95±0.42 & 45.93±1.04 & 45.68±0.95 & 17.96±0.76 & 33.38±1.60 & 34.13±1.39 \\
\moours{\limber}{(all)}{}{} & 39.86±0.30 & 136.31±0.63 & 23.85±0.81 & 83.74±2.80 & 22.17±3.39 & 75.51±19.68 & 73.42±0.08 & 72.73±0.10 & 47.98±1.87 & 47.70±1.58 & 31.25±0.50 & 36.19±2.42 & 38.93±2.86 \\
\moours{\mapl}{}{}{} & 36.96±1.60 & 126.05±5.13 & 21.15±2.71 & 69.20±11.74 & 18.43±0.82 & 50.57±3.99 & 67.13±1.28 & 66.76±1.35 & 45.94±1.76 & 45.65±1.62 & 21.04±0.88 & 36.21±1.24 & 37.02±0.45 \\
\moours{\epalm}{}{}{} & 33.79±0.43 & 115.34±1.23 & 17.98±0.79 & 64.65±1.38 & 16.27±0.37 & 42.58±0.54 & 59.34±0.21 & 59.03±0.25 & 41.38±3.06 & 41.20±2.89 & 16.59±0.93 & 23.52±6.40 & 27.82±1.59 \\
\midrule
\suniLzero{}{} & 38.66±1.25 & 131.29±3.38 & 25.09±0.37 & 87.05±1.61 & 22.23±0.91 & 73.67±5.36 & 70.11±0.14 & 69.56±0.19 & 48.25±1.20 & 47.80±1.17 & 22.97±1.22 & 37.69±0.65 & 38.45±1.48 \\
\suniInner{}{} & 38.80±0.66 & 130.91±1.04 & 21.34±0.38 & 75.86±0.92 & 21.21±0.30 & 65.22±1.55 & 67.88±0.28 & 67.64±0.11 & 45.27±0.38 & 44.92±0.70 & 23.70±0.96 & 35.98±0.68 & 36.36±1.38 \\
\suniRavgpool{}{} & 38.52±0.88 & 131.77±3.50 & 24.68±1.35 & 86.09±4.45 & 20.04±0.86 & 61.18±4.87 & 64.84±2.07 & 64.61±2.14 & 48.86±0.76 & 48.56±0.90 & 19.14±0.60 & 35.17±1.34 & 35.41±1.78 \\
\suniRlinear{}{} & 38.97±0.32 & 133.01±0.61 & 24.69±1.74 & 85.31±5.26 & 21.11±0.93 & 69.76±2.77 & 64.76±0.27 & 64.45±0.31 & 47.66±0.52 & 47.31±0.31 & 19.08±0.93 & 34.58±0.72 & 35.30±1.56 \\
\suniRqsformer{}{} & 38.62±1.72 & 131.92±5.41 & 22.76±1.17 & 75.46±3.95 & 17.50±0.59 & 51.03±2.04 & 61.09±2.27 & 60.89±2.20 & 46.08±0.30 & 45.91±0.17 & 18.56±0.71 & 35.35±0.58 & 35.63±0.78 \\
\suniRrand{}{} & 39.63±0.48 & 134.90±0.67 & 24.60±0.53 & 86.84±1.49 & 19.49±1.33 & 58.15±5.21 & 71.33±0.09 & 70.76±0.10 & 47.60±0.96 & 47.25±1.08 & 21.28±1.59 & 35.00±1.29 & 34.74±1.34 \\
\sunicattn{}{} & 38.16±0.41 & 130.05±1.01 & 23.83±0.73 & 81.38±2.52 & -- & -- & 69.45±1.17 & 69.05±1.01 & 41.73±0.83 & 41.48±0.61 & -- & -- & -- \\

\bottomrule
\end{tabular}
}
\caption{
    Comparison of our proposed DePALM architectures and our baseline re-implementations, after training on 100\% or 1\% of each image dataset. We show the average score and standard deviation over five runs, in the format avg±std.
}
\label{tab:image_results}
\end{table*}

\paragraph{Additional experiments.}
We provide complementary results that extend the ones in Table 4 of the main paper.
In \Cref{tab:image_results} we add standard deviation across the five runs on image datasets, and add additional metrics (BLEU for COCO and TextCaps, test set performance for VQAv2).
In \Cref{tab:audio_results} and \Cref{tab:video_results}, we do similarly but for AudioCaps (audio captioning) and MSRVTT (video captioning).

In addition, in \Cref{tab:image_results_dino}, we report the results on image datasets using the DINOv2 visual encoder in place of the CLIP models for the COCO and VQAv2 benchmarks. We see that the scores slightly degrade, compared to \Cref{tab:image_results}, but still better than previous state-of-the-art parameter-efficient results. In particular, with the larger dataset (VQAv2), the effect is smaller, especially when using \suniLzero{}{}. This reveals that models trained in an unsupervised setting could be a good candidate to adapt LLMs efficiently to do multimodal tasks.

\begin{table}[t]
\scriptsize
\centering	
\setlength\tabcolsep{4pt}
\resizebox{\linewidth}{!}{%
\begin{tabular}{lcccccc}
\toprule	 	
Method (on AudioCaps)
    & B@1 & B@2 & METEOR & CIDEr & SPICE & SPIDER
    \\

\moours{\limber}{(1)}{}{} & 69.38±0.70 & 51.11±0.43 & 21.90±0.14 & 62.04±0.82 & 15.84±0.28 & 38.94±0.47 \\
\moours{\limber}{(all)}{}{} & 69.52±0.69 & 51.23±0.74 & 22.53±0.30 & 64.34±2.44 & 15.91±0.65 & 40.12±1.37 \\
\moours{\mapl}{}{}{} & 70.05±1.27 & 52.11±1.10 & 22.87±0.20 & 65.36±1.61 & 16.42±0.34 & 40.89±0.82 \\
\moours{\epalm}{}{}{} & 61.94±2.01 & 45.83±1.75 & 21.38±0.46 & 60.84±2.31 & 15.41±0.53 & 38.13±1.37 \\
\midrule
\suniLzero{}{} & 71.54±0.89 & 53.37±1.05 & 23.66±0.33 & 69.70±2.31 & 17.03±0.63 & 43.37±1.42 \\
\suniInner{}{} & 70.85±1.63 & 52.75±1.87 & 23.12±0.48 & 65.96±3.55 & 16.44±0.71 & 41.20±2.09 \\
\suniRavgpool{}{} & 68.91±1.02 & 50.81±1.44 & 23.07±0.39 & 66.80±3.69 & 16.29±0.34 & 41.54±1.81 \\
\suniRlinear{}{} & 68.73±1.46 & 50.59±1.15 & 23.08±0.35 & 65.52±2.89 & 16.32±0.28 & 40.92±1.42 \\
\suniRqsformer{}{} & 69.89±0.88 & 51.82±0.72 & 22.69±0.28 & 66.16±2.38 & 16.19±0.32 & 41.17±1.17 \\
\suniRrand{}{} & 70.44±0.64 & 52.22±1.13 & 22.99±0.43 & 66.38±3.38 & 16.36±0.48 & 41.37±1.84 \\

\bottomrule
\end{tabular}
}
\caption{
    Comparison of our proposed DePALM architectures and our baseline re-implementations, after training on AudioCaps. We show the average score and standard deviation over five runs, in the format avg±std.
}
\label{tab:audio_results}
\end{table}
\begin{table}
\scriptsize
\centering	
\setlength\tabcolsep{4pt}
{
\begin{tabular}{lccc}
\toprule	 	
Method (on AudioCaps)
    & B@4 & METEOR & CIDEr
    \\

\moours{\limber}{(1)}{}{} & 34.22±1.04 & 27.56±0.35 & 46.03±2.11 \\
\moours{\limber}{(all)}{}{} & 36.30±0.87 & 27.77±0.36 & 46.87±1.80 \\
\moours{\mapl}{}{}{} & 36.78±1.22 & 28.01±0.24 & 47.27±1.90 \\
\moours{\epalm}{}{}{} & 25.59±1.28 & 25.35±0.35 & 38.83±2.14 \\
\midrule
\suniLzero{}{} & 38.78±1.51 & 28.54±0.37 & 49.88±2.01 \\
\suniInner{}{} & 39.44±1.07 & 28.29±0.47 & 47.76±2.18 \\
\suniRavgpool{}{} & 39.36±1.51 & 28.59±0.29 & 50.52±2.24 \\
\suniRlinear{}{} & 40.56±1.23 & 28.71±0.36 & 51.60±2.28 \\
\suniRqsformer{}{} & 38.32±1.05 & 27.41±0.32 & 45.49±1.49 \\
\suniRrand{}{} & 36.39±0.88 & 27.85±0.47 & 47.90±2.30 \\

\bottomrule
\end{tabular}
}
\caption{
    Comparison of our proposed DePALM architectures and our baseline re-implementations, after training on MSRVTT for captioning. We show the average score and standard deviation over five runs, in the format avg±std.
}
\label{tab:video_results}
\end{table}
\begin{table*}
\small
\centering	
\setlength\tabcolsep{4pt}
\resizebox{\linewidth}{!}{%
\begin{tabular}{lcc<{\hspace{8pt}}cc<{\hspace{8pt}}cc<{\hspace{8pt}}cc}
\toprule	 	
\multirow{2}{*}{Method}
    & \multicolumn{2}{c<{\hspace{8pt}}}{COCO}
    & \multicolumn{2}{c<{\hspace{8pt}}}{COCO (1\% data)}
    & \multicolumn{2}{c<{\hspace{8pt}}}{VQAv2}
    & \multicolumn{2}{c<{\hspace{8pt}}}{VQAv2 (1\% data)}
    \\
\cmidrule(lr{8pt}){2-3} \cmidrule(lr{8pt}){4-5} \cmidrule(lr{8pt}){6-7}  \cmidrule(lr{8pt}){8-9}
    & B@4 & CIDEr
    & B@4 & CIDEr
    & Val & Test
    & Val & Test
    \\

\midrule

\moours{\limber}{(1)}{}{} & 31.28±0.62 & 106.93±1.35 & 16.55±0.55 & 61.54±1.55 & 55.63±0.07 & 55.30±0.19 & 20.01±17.72 & 19.79±17.61 \\
\moours{\limber}{(all)}{}{} & 37.86±0.28 & 129.24±0.75 & 22.21±0.99 & 74.96±4.21 & 68.95±2.42 & 68.64±2.42 & 44.56±0.46 & 44.31±0.52 \\
\moours{\mapl}{}{}{} & 36.22±1.55 & 122.28±4.99 & 21.39±1.09 & 70.50±2.42 & 66.37±0.22 & 66.13±0.23 & 45.47±0.46 & 45.38±0.30 \\
\moours{\epalm}{}{}{} & 31.28±0.24 & 106.21±0.56 & 16.34±0.68 & 57.45±1.38 & 57.62±0.16 & 57.37±0.25 & 39.90±1.32 & 39.41±1.64 \\
\midrule
\suniLzero{}{} & 37.68±0.39 & 127.38±1.11 & 23.29±0.74 & 79.45±2.22 & 68.42±0.10 & 68.05±0.12 & 47.91±0.91 & 47.51±0.92 \\
\suniInner{}{} & 37.28±0.61 & 124.53±1.28 & 22.27±0.34 & 73.53±1.20 & 65.73±0.25 & 65.49±0.27 & 43.81±0.50 & 43.52±0.40 \\
\suniRavgpool{}{} & 37.24±0.45 & 126.29±0.96 & 22.52±0.53 & 77.53±1.82 & 63.63±2.14 & 63.66±2.03 & 45.51±0.51 & 45.08±0.54 \\
\suniRlinear{}{} & 37.14±0.33 & 125.29±0.67 & 22.37±0.43 & 76.71±2.13 & 66.47±0.20 & 66.48±0.22 & 45.62±0.49 & 45.29±0.45 \\
\suniRqsformer{}{} & 36.00±1.59 & 122.30±5.36 & 20.74±1.49 & 65.85±5.08 & 59.93±0.30 & 59.79±0.55 & 48.09±0.46 & 47.64±0.35 \\
\suniRrand{}{} & 37.13±0.59 & 126.44±1.20 & 23.02±0.41 & 79.68±1.16 & 63.24±2.07 & 63.10±1.95 & 44.23±0.50 & 44.07±0.42 \\
\sunicattn{}{} & 36.41±0.42 & 122.69±1.02 & 8.34±0.83 & 18.03±4.04 & 44.84 & 44.89 & -- & -- \\

\bottomrule
\end{tabular}
}
\caption{
    Comparison of our proposed DePALM architectures and our baseline re-implementations, after training on 100\% or 1\% of COCO and VQAv2 datasets, when using the DINOv2 as the perceptual backbone extractor. We show the average score and standard deviation over five runs, in the format avg±std.
}
\label{tab:image_results_dino}
\end{table*}
\begin{table}
\scriptsize
\centering	
\setlength\tabcolsep{4pt}
{
\begin{tabular}{lcccc}
\toprule	 	
\multirow{2}{*}{Method}
    & COCO
    & TextCaps
    & AudioCaps
    & MSRVTT
    \\
\cmidrule(lr{8pt}){2-2} \cmidrule(lr{8pt}){3-3} \cmidrule(lr{8pt}){4-4}  \cmidrule(lr{8pt}){5-5}
    & CIDEr
    & CIDEr
    & SPIDEr
    & CIDEr
    \\

\midrule

\moours{\limber}{(all)}{}{}
    & 136.31        
    & 75.51        
    & 40.12             
    & 46.87            
    \\

\suniLzero{}{}
    & 131.29        
    & 73.67        
    & 43.37             
    & 49.88             
    \\

\midrule

\moours{\limber}{(all)}{}{} + bias-FT
    & 137.37      
    & 74.12       
    & 45.45      
    & 49.16      
    \\

\suniLzero{}{} + bias-FT
    & 133.55      
    & 67.98      
    & 47.35     
    & 50.86      
    \\

\bottomrule
\end{tabular}
}
\caption{
Impact of bias fine-tuning in the perceptual backbone  model. The results are averaged over 5 runs.
}
\label{tab:bias_ft}
\end{table}

\paragraph{Efficient fine-tuning of the feature model.}
In our experiments so far, we  used prompt-tuning, but did not fine-tune any internal parameters of the LLM or feature backbone.
In \Cref{tab:bias_ft} we consider the impact of adding bias-tuning to the feature model. 
While adding only 0.5M learnable parameters, we observe substantial gains on COCO,  AudioCaps, and MSRVTT, but surprisingly  observed  performance loss on smaller datasets such as TextCaps. So this method should mostly be considered given enough data.
For simplicity, and to keep good results on small datasets, we used only prompt tuning for the LLM and kept the encoders completely frozen in all other experiments.

\section{Carbon footprint estimation} 
\label{app:footprint}

We report the estimated carbon footprint of training a \textit{single} instance of \suniLzero{}{} for four different datasets, using the following method. 
We take the average training time $T$, and then compute the total GPU hours $T_{\text{GPU}}=T\times 8$, as we use a single 8-GPU node for each model.
We then estimate the power consumption in kWh, given a Thermal Design Power (TDP) of the V100-32G GPU equal to 250W and a Power Usage Effectiveness (PUE) of 1.1,  as $K=\frac{250 \times 1.1}{1000} \times T_{\text{GPU}}$. 
Finally, given a carbon intensity factor of 0.385 kg CO$_2$ per KWh, we obtain the emission $E$ in kg of CO$_2$ as $E=0.385 \times K$.  

\begin{table}[h]
\centering	
\setlength\tabcolsep{4pt}
{\scriptsize
\begin{tabular}{lcccc}
\toprule
 & COCO & OKVQA & AudioCaps & MSRVTT \\
\midrule
Training time: $T$ & 2h14 & 1h23 & 2h21 & 0h31 \\
GPU hours (8 GPUs): $T_\textrm{GPU}$ & 17.87 & 11.07 & 18.80 & 4.13 \\
Estimated kWh: $K$ & 4.91 & 3.04 & 5.17 & 1.14 \\
Emitted kg of CO$_2$: $E$ & 1.89 & 1.17 & 1.99 & 0.44 \\
\bottomrule
\end{tabular}
}
\caption{Estimated carbon footprint of training a single DePALM\textsuperscript{QP,L0} model, on four different datasets.}
\label{tab:carbon}
\end{table}

\end{document}